\documentclass{iopjournal}

\usepackage{graphicx}%
\usepackage{multirow}%
\usepackage{amsmath,amssymb,amsfonts}%
\usepackage{amsthm}%
\usepackage{mathrsfs}%
\usepackage[title]{appendix}%
\usepackage{xcolor}%
\usepackage{textcomp}%
\usepackage{manyfoot}%
\usepackage{booktabs}%
\usepackage{algorithm}%
\usepackage{algorithmicx}%
\usepackage{algpseudocode}%
\usepackage{listings}%

\usepackage[utf8]{inputenc} 
\usepackage[T1]{fontenc}    
\usepackage{xcolor}         
\definecolor{citeblue}{rgb}{0.41,0.69,0.84}

\usepackage{url}            
\usepackage{booktabs}       
\usepackage{amsfonts}       
\usepackage{nicefrac}       
\usepackage{microtype}      

\usepackage{optidef}
\usepackage{amsmath}
\usepackage{amssymb}
\usepackage{amsthm}
\usepackage{amssymb}
\usepackage{multirow}
\usepackage{multicol}
\usepackage{graphicx}
\usepackage{subcaption}
\usepackage{fancyhdr}
\usepackage{newfloat}
\usepackage{listings}

\theoremstyle{definition}
\newtheorem{definition}{Definition}[section]
\theoremstyle{plain}
\newtheorem{theorem}{Theorem}[section]
\newcommand{\Hc}{\ensuremath{\mathcal{H}}}
\newcommand{\Lc}{\ensuremath{\mathcal{L}}}
\newcommand{\Nc}{\ensuremath{\mathcal{N}}}

\newcommand{\bq}{\mathbf{q}}
\newcommand{\bp}{\mathbf{p}}
\newcommand{\by}{\mathbf{y}}

\newcommand{\Jc}{\ensuremath{\mathcal{J}}}
\newcommand{\RR}{\ensuremath{\mathbb{R}}}

\DeclareCaptionStyle{ruled}
{labelfont=normalfont,labelsep=colon,strut=off} 
\floatstyle{ruled}
\newfloat{listing}{tb}{lst}{}
\floatname{listing}{Listing}
\theoremstyle{thmstyleone}%
\theoremstyle{thmstyletwo}%

\theoremstyle{thmstylethree}%

\usepackage{xcolor}

\begin{document}

\articletype{Paper} 

\title{Symplectic Neural Networks for learning Non-Separable Hamiltonians}

\author{Harsh Choudhary$^{1,*}$\orcid{0009-0000-6467-3114}, Vyacheslav Kungurtsev$^1$\orcid{0000-0003-2229-8824}, Chandan Gupta$^2$\orcid{0000-0002-2371-3586} Melvin Leok$^3$\orcid{0000-0002-8326-0830} and Georgios Korpas $^{1,4}$\orcid{0000-0003-3850-4979}}

\affil{$^1$Computer Science Department, Czech Technical University in Prague, Prague 2, Czechia 12000}

\affil{$^2$Computer Science Department, Indraprasth Institute of Information Technology, New Delhi, India  110020}

\affil{$^3$Department of Mathematics, University of California at San Diego, La Jolla, CA 92093}


\affil{$^4$	Quantum Technologies Group, HSBC, Singapore, 117439, Singapore}

\affil{$^*$Computer Science Department, Czech Technical University in Prague, Prague 2, Czechia 12000}

\email{choudhar@fel.cvut.cz}


\begin{abstract}
Hamiltonian Neural Networks (HNNs) integrate physical priors into neural models by learning a system's Hamiltonian, improving generalization and sample efficiency.  Identifying the system Hamiltonian from noisy observations of state variables is a challenging task. For simulations to faithfully reflect the long-term behavior of Hamiltonian systems, especially energy conservation, it is essential to use symplectic integrators, which preserve the system’s geometric structure. This fidelity comes at a cost: implicit symplectic integrators are more computationally intensive and make backpropagation through the ODE solver non-trivial. However, by leveraging the fact that symplectic discretizations of the adjoint system yield the same sensitivities associated by backpropagation, we obtain an efficient method of training the Neural Network parameters. In our work, we explore this alternate method of HNN training under noisy observation of trajectories with our HNN model based on an implicit symplectic integrator. Computationally, a predictor-corrector based ODE solver and fixed point iteration help to mitigate the computational cost of the implicit timestepping, resulting in more efficient generation of gradient updates. We showcase the numerical advantage, in experiments, in system identification and energy preservation on a range of non-separable, chaotic systems and the efficient computation and memory complexity of our method. We also observe that the post-processing of the learned Hamiltonian using backward error analysis yields a modified Hamiltonian that is a more accurate approximation of the true Hamiltonian without the need to use  more accurate discretizations of the flow map.
\end{abstract}

\section{Introduction}\label{sec:intro}
\medskip
The Hamiltonian formalism is standard in mathematical physics for expressing dynamics in many physical systems. Beyond yielding a systematic framework for deriving the dynamics of the system, there are deep structural, geometric, topological, and analytic properties of Hamiltonian functions that can be used to understand the physics of the system and its dynamics~\cite{MaRa1999, kang1991hamiltonian}. Hamiltonian systems are described by a single scalar function in the phase space $\Hc : \Omega \subseteq \RR^{2d} \rightarrow \RR$ and the corresponding system for $y \in \Omega$ in Hamilton's equations~\cite{hand1998analytical, SanzSerna2018}.

\begin{equation}\label{eq:Hamilt_ODE}
\begin{aligned}
\dot{y} = {J}^{-1}\nabla \Hc(y), \quad {J} = \begin{pmatrix}
0 & 1\\
-1 & 0
\end{pmatrix},
\end{aligned}
\end{equation}
where $J$ is the canonical symplectic matrix, which we will refer to repeatedly in subsequent sections. The solutions of the system are trajectories $y(t)$. If the Hamiltonian does not have an explicit time-dependence, then it is the Noether quantity associated with time-translational symmetry and is conserved along the flow, i.e., $\Hc(y(t))$ remains constant along trajectories. This implies many important conservation properties across physical systems of interest.
For a set of canonical coordinates \{$p_i$, $q_i$\} $\in \RR^{2d}$,  an integrator, or numerical simulation method, is said to be symplectic if it preserves the canonical symplectic form, $\omega = \sum_{i=1}^{n} dp_i \wedge dq^i$; see subsequent sections and also~\cite{channell1990symplectic}.

Geometric integrators preserve geometric invariants of the flow, and the use of such integrators has been instrumental in understanding the long-term qualitative properties of many important physical systems \cite{Ruth1983,HaLuWa2006,FeQi2010, maslovskaya2024symplectic, valperga2022learning}. 
As such, by learning the Hamiltonian itself rather than seeking to learn trajectories, we are able to directly learn the qualitative properties of the system. In contrast, non-geometric numerical methods such as the Euler and explicit Runge--Kutta methods do not conserve the Hamiltonian \cite{HaLuWa2006}. A result in \cite{GeMa1988} shows that it is not possible for a fixed-timestep integrator to simultaneously preserve the energy and the symplectic structure unless it samples the exact solution, and this result inspired the development of energy-preserving and symplectic integration schemes. 

In recent years, the field of Physics-Informed Machine Learning has brought significant advances in architecture and learning techniques. By enforcing physical inductive biases while learning the dynamics through appropriate structural modifications to standard NN training, the learning process requires much fewer samples to accurately fit the data and achieves better out-of-distribution accuracy. This includes the recent development of Hamiltonian Neural Networks \cite{bertalan2019learning}. By representing the Hamiltonian as a neural network, a more physical representation of the system is available. While incorporating symplectic integrators yields more qualitatively accurate approximations of the Hamiltonian dynamics, it poses technical challenges, namely (i) reconstructing a learned Hamiltonian function from noisy trajectory observations, (ii) solving the implicit equations required by symplectic integration methods, and (iii) ensuring that the learned Hamiltonian model generalizes to out-of-distribution data. Motivated by these technical challenges, the remainder of this paper is structured as follows:

In Section \ref{sec:background}, we provide background on two complementary formulations of dynamical systems: the Lagrangian and Hamiltonian perspectives. Section \ref{s:symp_int} reviews symplectic integrators and their application to solving general Hamiltonian systems. We also discuss briefly backward error analysis in the context of symplectic methods. In Section \ref{sec:method}, we describe the architecture of our Hamiltonian Neural Network and introduce an implicit symplectic integrator that is implemented with a predictor-corrector method in the forward pass. Section \ref{ss:adjoint_sens} describes the backward pass and introduces the adjoint sensitivity equations for Hamiltonian Learning. Finally, Section \ref{sec:numerics} presents numerical experiments. 

\subsection{Contributions}
\medskip
The paper makes the following contributions:
\begin{itemize}
    \item Showing that a neural framework can be constructed which learns non-separable Hamiltonians up to some error tolerance from noisy observations of trajectories without having any implicit structural bias (no assumptions on the separability of the Hamiltonian).
    \item Showing that the Hamiltonian Neural Net generalizes to out-of-distribution data and is able to learn the governing Hamiltonians for systems with chaotic dynamics.
    \item Using the fact that the adjoint sensitivity of a symplectic discretization of a Hamiltonian ODE can be computed using the same symplectic discretization applied to the adjoint system, we are able to compute exact gradients of the loss with respect to neural network parameters without the need for backpropagation.
    \item We also describe in brief how backward error analysis applied to our Learned Hamiltonian results in a more accurate approximation of the true Hamiltonian.
\end{itemize}

\section{Hamiltonian dynamics: A background}\label{sec:background}
\medskip
Before discussing the Hamiltonian formulations, we recall the definition of a manifold in the context of mechanics. 

\begin{definition}
    A \emph{manifold} $Q$ is a space that is locally Euclidean and it serves as the geometric setting for describing the state of a system. The \emph{configuration space} of a mechanical system is modeled as a smooth manifold $Q$, where each point  $q \in Q$  represents a possible configuration of the system.

Associated with $Q$ are two fundamental constructions:
\begin{itemize}
    \item The \emph{tangent bundle} \( TQ \), whose elements are pairs \( (q, \dot{q}) \), representing positions and velocities. This is the natural domain of the Lagrangian formalism.
    \item The \emph{cotangent bundle} \( T^*Q \), whose elements are pairs \( (q, p) \), representing positions and momenta. This serves as the phase space in Hamiltonian mechanics.
\end{itemize}

These bundles are themselves smooth manifolds of dimension \( 2d \), where \( d = \dim Q \). The dynamics of mechanical systems are described by differential equations on these manifolds, and respect their geometric structure.
\end{definition}

There are two equivalent and foundational approaches to formulating classical mechanics, rooted in geometric and variational principles: the Lagrangian and Hamiltonian frameworks. These frameworks describe the evolution of physical systems using generalized coordinates and velocities or momenta, rather than relying directly on Newton’s second law. The Lagrangian formalism is based on the principle of stationary action and encodes the dynamics in terms of the Lagrangian, which is difference between kinetic and potential energy. The Hamiltonian formalism reformulates the problem as a system of first-order differential equations on phase space, where the state of the system is expressed through canonical coordinates and conjugate momenta. These approaches systematically reveal conserved quantities through symmetries, as formalized by Noether’s theorem, and reveals the mathematical structure underlying the dynamics, including the preservation of the symplectic geometric structure in the Hamiltonian setting. Beyond their classical role, these formulations have become central in the analysis of dynamical systems, numerical methods, and emerging applications in optimization and machine learning.

\subsection{The Lagrangian formulation}
\medskip
Lagrangian mechanics describes the motion in a mechanical system by means of the configuration space. The configuration space of a mechanical system has the structure of a differentiable manifold, on which its group of diffeomorphisms acts. The basic ideas and theorems of Lagrangian mechanics are invariant under the right action of this group~\cite{arnol2013mathematical}, which reflect the coordinate independence of this formulation. A Lagrangian mechanical system is given by a configuration manifold and a Lagrangian function on its tangent bundle. Every one-parameter group of diffeomorphisms of the configuration space induces a tangent lifted action on the tangent bundle. If the Lagrangian is invariant under this tangent lifted action, then there is an associated Noether quantity that is conserved. A Newtonian potential system is a particular case of a Lagrangian system where the configuration space is Euclidean and the Lagrangian function is the difference between kinetic and potential energies. The equations of motion are derived via the variational principle by extremizing the action functional \eqref{eq:action} whose domain is the infinite-dimensional space of curves and the dynamics is defined by the curve that extremizes this action\footnote{The action functional is extremized when the first variation vanishes; it may correspond to a minimum, maximum, or saddle point.}.

Let $Q$ be a $d$-dimensional smooth manifold representing the configuration space of a mechanical system. The tangent bundle $TQ$ of $Q$ consists of all pairs $(q, \dot{q})$, where $q \in Q$ denotes a configuration and $\dot{q} \in T_q Q$ is a tangent vector representing the velocity at $q$. Given local coordinates $(q_1, \ldots, q_d)$ on $Q$, the tangent bundle has local coordinates $(q_i, \dot{q}_i)$ for $i = 1, \ldots, d$. Lagrangian mechanics is formulated on $TQ$ with the action functional $S: C^2([t_0,t_1],Q) \to \mathbb{R}$ is defined as:
\begin{equation}\label{eq:action}
\begin{aligned}
S[q] = \int_{t_0}^{t_1} L(t, q(t), \dot{q}(t)) dt.
\end{aligned}
\end{equation}
Then, the dynamics of the system is governed by Hamilton's principle:
$$
\delta S[q] = 0,
$$
for all variations $\delta q(t)$ of $q(t)$ that vanish at the endpoints, i.e., $\delta q(t_0) = \delta q(t_1) = 0$. Computing the variation, integrating by parts, and evoking the fundamental theorem of the calculus of variations yields the Euler--Lagrange equations:
\begin{align}\label{eq:lagrange}
\frac{d}{dt} \left( \frac{\partial L}{\partial \dot{q}_i} \right) - \frac{\partial L}{\partial q_i} = 0, \quad \text{for } i = 1, \ldots, d.
\end{align}
\subsection{The Hamiltonian formulation}
\medskip
Another approach is the Hamiltonian formulation of dynamics, which provides an alternative (but equivalent) description of the system on phase space when the Lagrangian is hyperregular. In this picture, one works on the cotangent bundle $T^*Q$ (the space of pairs $(q,p)$ of generalized coordinates and conjugate momenta) instead of the tangent bundle $TQ$ consisting of $(q,\dot q)$ used in Lagrangian mechanics. The cotangent bundle $T^*Q$ carries a natural symplectic structure, which is central to Hamiltonian mechanics. Specifically, let $\mathcal{M} = T^*Q$ be a $2d$-dimensional smooth manifold. Then the symplectic form $\omega$ is a closed, non-degenerate 2-form on $\mathcal{M}$, and in canonical (Darboux) coordinates $(q_1, \ldots, q_d, p_1, \ldots, p_d)$ on $T^*Q$, it is given by:
\begin{align}\label{eq:symplecticform}
    \omega = \sum_{i=1}^d dq_i \wedge dp_i.
\end{align}

To transition from the Lagrangian $L: TQ \to \mathbb{R}$ to the Hamiltonian formalism, we first define the conjugate momenta by taking partial derivatives of $L$ with respect to the generalized velocities. In local coordinates $q_i$\footnote{Strictly speaking, \( q_i \) should be written as \( q^i \), since it denotes a component of a vector, while \( p_i \) is a component of a covector. However, for notational simplicity and because we work primarily with individual components when formulating our loss functionals, we use subscripts for both.} on $Q$ , the $i$-th component of momentum is defined as:

\begin{align*}
    p_i = \frac{\partial L}{\partial \dot{q_i}}, \quad \text{for } i=1,\dots,d 
\end{align*}
This definition induces a map known as the Legendre transform, often denoted $\mathbb{F}L: TQ \to T^*Q$. The Legendre transform sends a point $(q,\dot q)$ in the tangent bundle to a corresponding point $(q,p)$ in the cotangent bundle by pairing velocities with momenta. In coordinates, $\mathbb{F}L(q,\dot q) = (q,p)$, where $p_i = \partial L/\partial \dot q_i$\footnote{We assume $L$ is regular, meaning the Hessian matrix $\big(\partial^2 L/\partial \dot q_i \partial \dot q_j\big)$ is nonsingular, so that this transformation is locally invertible. If this induces a global diffeomorphism $TQ \cong T^*Q$, then the Lagrangian is said to be hyperregular. Under this assumption, for each $(q,p)$, there is a unique $\dot q$ such that $p_i = \partial L/\partial \dot q_i$.}.

Given this correspondence, we define the Hamiltonian $H: T^*Q \to \mathbb{R}$ as the Legendre transform of the Lagrangian, i.e., the function whose value is the “energy” obtained by trading the velocity dependence of $L$ for momentum dependence. It is given by the formula,
\begin{equation}
H(q, p) = \sum_{i=1}^d p_i \dot{q}_i - L(q, \dot{q}),\label{H_in_terms_of_L}
\end{equation}
where $\dot{q}_i$ is expressed as a function of $(q, p)$ by inverting the Legendre transform. Equivalently, the dependence on the velocities on the right-hand side can be eliminated by extremizing with respect to the velocities, which is analogous to the approach adopted in the Pontryagin maximum principle. Given $H$, we can define a unique vector field $X_H$ on $T^*Q$, the Hamiltonian vector field, by the condition
\begin{align}\label{Hamilton_eq_abstract}
    dH = \omega(X_H,\cdot).
\end{align}
The Hamiltonian vector field $X_H$ takes the form 
\begin{align}
    X_H = \sum_{i=1}^{d} \frac{\partial H}{\partial p_i}\frac{\partial }{\partial q_i} - \frac{\partial H}{\partial q_i} \frac{\partial}{\partial p_i}.
\end{align}
The integral curves of $X_H$ are the solutions to Hamilton's equations 
$$
\dot{p}_i = -\frac{\partial H}{\partial q_i}, \quad \dot{q}_i = \frac{\partial H}{\partial p_i}, \quad \text{for } i = 1, \ldots, d.
$$
While we have derived Hamilton's equations using local canonical coordinates, \eqref{Hamilton_eq_abstract} defines a global vector field on $\mathcal{M}$.
These equations describe the flow of the system in phase space and are equivalent to the Euler--Lagrange equations \eqref{eq:lagrange} derived from Hamilton's principle if the Legendre transformation is globally invertible and the Lagrangian $L$ is related to the Hamiltonian $H$ by \eqref{H_in_terms_of_L}.

For completeness, the Hamilton--Jacobi partial differential equation,
$$
\partial_t S + H(q, \partial_q S) = 0,
$$
describes the generating function $S$ for the canonical transformation that maps $(q(t_0),p(t_0))$ to $(q(t_1),p(t_1))$. This provides an alternative method for solving Hamilton's equations \cite{AbMa1978}, and Jacobi's solution to the Hamilton--Jacobi equation is given in terms of the action functional evaluated along the two-point boundary value solution of the Euler--Lagrange equations.

 \section{Symplectic integrators}\label{s:symp_int}
 \vspace{0.5em}


This section presents symplectic integration in the context of the current Hamiltonian learning literature. Then, we will review some properties of symplectic flow maps which motivate their incorporation into the learning process for dynamical systems. Finally, we present some standard methods for constructing symplectic integrators that we later use in this work. 

\subsection{Symplectic integration in NNs: Previous work}\label{ss:prev_work}
\medskip

Since the original proposal by \cite{greydanus2019hamiltonian} and the concurrent work by \cite{bertalan2019learning}, HNNs have generated much scientific interest. This has spawned generative 
\cite{toth2019hamiltonian}, recurrent 
\cite{chen2019symplectic}, and constrained 
\cite{zhong2019symplectic}
versions, as well as Lagrangian Neural Networks 
\cite{cranmer2020lagrangian}
have been proposed.  
In contrast to standard HNNs, the approach introduced in \cite{chen2019symplectic} directly optimizes the actual states observed at each timestep for a given initialization by integrating the Hamiltonian vector field using a symplectic integrator (leapfrog algorithm) and backpropagating each squared error through time. The state at the next timestep is predicted using the symplectic integrator; in this way the entire time series is predicted, which is then compared with the observed data. Note that they make the assumption that the Hamiltonian is separable, which is significant as the leapfrog algorithm is generally implicit, but if the Hamiltonian is separable, then the algorithm becomes explicit. Notably, the work in \cite{vsipka2023direct} uses an implicit midpoint scheme to train a Poisson Neural Net, which learns the Poisson structure of a dynamical system, and the network is trained using a standard backpropagation technique.

Most of these works considered either explicit integration schemes, some semi-implicit schemes, or used a separable ansatz for the neural nets. In \cite{khoo2024separable, Wu2024}, additive separability biases were introduced in the HNN architecture/training, allowing the network to learn Hamiltonians of the form $H(q,p)=T(p)+V(q)+\text{const}$. This yields better performance by making the Störmer–Verlet (leapfrog) integrators fully explicit and easier to train, at least when the true Hamiltonian is well-approximated by a separable Hamiltonian. The error bounds for such symplectic HNNs have been analyzed for the noiseless case in recent works (e.g., showing that energy error grows linearly in time under a symplectic integrator) \cite{canizares2024hamiltonian, david2023symplectic}, and it was established that the learned $H$ is reasonable but is slightly perturbed from the true $H$. These limitations highlight that while semi-implicit methods make training feasible, they come at the cost of a small modeling bias as the integrator inherently assumes that the Hamiltonian is separable in order to simplify the computation. More details can be found in \cite{hairer2006geometric, yoshida1990construction}.

A generalized HNN framework is proposed in \cite{xiong2020nonseparable}, which can be used for non-separable Hamiltonians where they approximate the original non-separable Hamiltonian by an augmented one that was introduced in \cite{tao2016explicit}, with an extended phase space and a tunable parameter $\omega$ which controls the binding between the two copies of the Hamiltonian and model them using NNs. The approach in \cite{tao2016explicit} yields an explicit symplectic integration scheme for non-separable Hamiltonians, at the expense of increased complexity associated with two coupled Hamiltonian systems, and the need to tune the coupling parameter. As we mentioned earlier, all such Symplectic HNN architectures learn a valid Hamiltonian, which is a perturbation of the true Hamiltonian, due to the approximation error in replacing the exact flow map with a numerical integrator. Establishing the error bounds is necessary, and there have been a few classic works that derive these error bounds for symplectic integrators, notably \cite{hairer2006geometric, david2023symplectic}.

Most existing works have either relied on explicit methods or employed imperfect implicit schemes, often limited to noiseless settings. This is largely due to the computational cost of the nonlinear solvers, which can be addressed by using a predictor-corrector approach. The predictor is an explicit one-step method with the same order as the implicit symplectic integrator, and the corrector involves a few fixed-point iterations of the symplectic integrator. Another challenge is posed by backpropagation through implicit ODE solvers, which scales in memory with the simulation length and number of parameters. In \cite{Rahma2024}, a backpropagation-free framework using sampled neural networks is proposed for learning Hamiltonians, which is effective with rich trajectory data but assumes access to the true time derivatives, which is unrealistic in practice. Notably another approach that avoids solving a forward integral is in form of SympNets\cite{jin2020sympnets} which learn the neural net itself as a symplectic map but this comes with restrictions on the freedom of choosing an arbitrary stepsize during simulation. We will try to address some of these issues through our work.


If the Hamiltonian is separable, it has the form, 
\begin{equation}\label{eq:sep_H}
\begin{aligned}
    \Hc(q,p) = T(p) + V(q)
\end{aligned}
\end{equation}
and the Hamiltonian vector field can be written as 
\begin{equation}\label{eq:seperable_Hamiltonian_dynamics}
    \begin{aligned}
        \dot{q} &= \frac{\partial T(p)}{\partial p}\\
\dot{p} &= -\frac{\partial V(q)}{\partial q}
    \end{aligned}
\end{equation}
For separable Hamiltonians, one can then use the standard St\"{o}rmer-Verlet method, whose numerical update is given by
\begin{equation}\label{eq:sv_int}
    \begin{aligned}
        p_{n+\frac{1}{2}} &= p_n - \frac{h}{2}\nabla V(q_n), \\
        q_{n+1}&=q_n + h\nabla T(q_{n+\frac{1}{2}}),\\
        p_{n+1}&=p_{n+\frac{1}{2}} - \frac{h}{2}V(q_{n+1}).
        \end{aligned}
\end{equation}

The method is second-order accurate but is only defined when the Hamiltonian is separable and the Hamiltonian vector field is given by \eqref{eq:seperable_Hamiltonian_dynamics}. In the next part, we will describe a well-known second-order symplectic method that works for general (non-separable) Hamiltonians.

\subsection{Symplectic integrators for nonseparable Hamiltonians}\label{ss:non-sep_H}
\medskip
When simulating physical systems, it is desirable to consider geometric integrators that preserve the geometric properties of the flow. More specifically, for Hamiltonian systems, this property is the preservation of the symplectic $2$-form described in \eqref{eq:symplecticform}. An automatic consequence of preserving the symplectic form is that the numerical integrator preserves the phase space volume on the cotangent bundle $T^*Q$.

There are various kinds of symplectic methods, the simplest of which are the first-order symplectic Euler schemes, which we will not discuss here. In what follows, we call a Hamiltonian \emph{non-separable} if it cannot be written in the form \eqref{eq:sep_H}, that is, as the sum of a kinetic term depending only on the momenta and a potential term depending only on the positions. Consider the canonical Hamiltonian dynamics where we do not assume separability,
\begin{equation}\label{eq:generic_Hamiltonian_dyn}
\begin{aligned}
\dot{q} &= \frac{\partial \Hc}{\partial p},\\
\dot{p} &= -\frac{\partial \Hc}{\partial q}.
\end{aligned}
\end{equation}
Observe that we cannot use methods of the form \eqref{eq:sv_int}.    
For systems with stiff dynamics or strongly nonlinear Hamiltonians, the semi-implicit methods may suffer from poor accuracy or numerical instability. In such cases, a fully implicit method is often desirable. The implicit midpoint method is often used to integrate such stiff dynamics, and it is symplectic and thus automatically phase space volume-preserving. Additionally, we observe that the map is symmetric, i.e., $\Phi_h^{-1} = \Phi_{-h}$. For the Hamiltonian vector field, the implicit midpoint method is written as
\begin{equation}\label{eq:im_itegrator}
\begin{aligned}
    q_{n+1} &= q_n + h \nabla_p \mathcal{H}\Big(\frac{q_n + q_{n+1}}{2}, \frac{p_n + p_{n+1}}{2}\Big), \\
    p_{n+1} &= p_n - h \nabla_q \mathcal{H}\Big(\frac{q_n + q_{n+1}}{2},\frac{p_n + p_{n+1}}{2} \Big),
\end{aligned}
\end{equation}
This update scheme is second-order accuracy and it is fully implicit in both variables $q$ and $p$. Hence, it requires a nonlinear solver.
\subsection{The numerical advantage of using symplectic methods}\label{ss:advantage_symp}
\begin{figure}[h!]
    \centering
\includegraphics[width=0.5\linewidth]{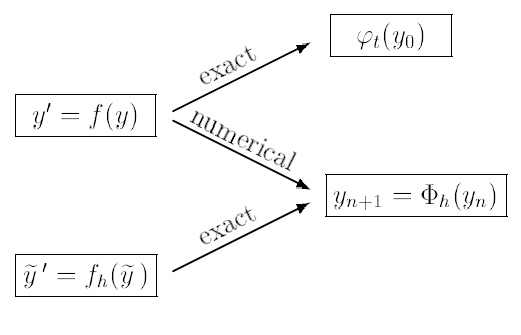}
    \caption{Original and modified ODE}
    \label{fig:backward_err}
\end{figure}

\medskip
Consider an initial value problem, 
\begin{align*}
    y' = f(y), \quad y(0) = y_0,
\end{align*}
and a numerical method $\Phi_h$ which approximates the trajectory with a sequence of points
$y_0, y_1,\dots, y_n$. Then, the forward error is concerned with the extent to which the numerical method approximates the exact flow,
\begin{align*}
    \epsilon = |y(nh) - \Phi_{nh}(y_0)|.
\end{align*}
In contrast, backward error analysis is concerned with constructing a modified differential equation $\tilde{y}' = f_h(\tilde{y})$ whose exact flow agrees with the numerical method. This relationship is expressed in Figure~\ref{fig:backward_err}, where $\varphi_t$ is the exact flow map for the ODE. It is in this context that the true power of symplectic integrators for simulating Hamiltonian dynamics is revealed. In general, a one-step method cannot be exactly expressed as the time-$h$ flow of a differential equation, but we can construct an asymptotic expansion for this modified equation,
\begin{equation}\label{eq:modified_eq}
\begin{aligned}
    \dot{\tilde{y}} = f_h(\tilde{y}) = f(\tilde{y}) + hf_1(\tilde{y}) + h^2f_2(\tilde{y})\dots ,\qquad \tilde{y}(0) = y_0
\end{aligned}
\end{equation}
Where $f_i$ denotes terms involving up to $i$-th derivatives of $f$.
If the numerical method is $p$-th order accurate, then the expansion $f_h$ agrees with $f$ up to the $h^{p-1}$ term, or that the first nontrivial correction term is $h^p$.
This backward error analysis can be applied to any numerical method, but when applied to a symplectic integrator, the modified vector field is Hamiltonian, which allows us to view a symplectic integrator as being very well approximated by the exact flow of a modified Hamiltonian. The precise theorem is as follows:
\begin{theorem}[\cite{hairer2003geometric}]\label{th:symplecticity}
    If a symplectic method $\Phi_h(y)$ is applied to a Hamiltonian system with a smooth Hamiltonian $\Hc:\RR^{2d}\rightarrow\RR$, then the modified equation \eqref{eq:modified_eq} is also Hamiltonian, more precisely, there exist smooth functions $\Hc_j:\RR^{2d}\rightarrow\RR$ for $j=2,3,\dots$ such that $f_j(y) = J^{-1}\Hc_j(y)$.
\end{theorem}
The theorem above implies that the numerical flow generated by the numerical map $\Phi$ is well approximated by the exact flow the modified Hamiltonian:
\begin{equation}\label{eq:modified_Hamiltonian}
\begin{aligned}
    \tilde{\Hc} = \Hc + h^{p}\Hc_{p}+\dots
\end{aligned}
\end{equation}
where $p$ denotes the order of accuracy of the numerical integrator and $\Hc_p$ satisfies $f_j=J^{-1}\Hc_j$ in the modified vector field. This is significant as symplectic integrators can be viewed as the flow of a modified Hamiltonian whereas the modified vector field associated with a non-symplectic integrator cannot be associated with a modified Hamiltonian flow. This ensures that the numerical reconstruction is more physical, as the total energy is approximately preserved over long time horizons.
Moreover, we will show in Section~\ref{sec:modified_Hamiltonian} an additional benefit this presents for learning physical systems. In essence, it allows us to use a lower-order symplectic integrator to learn a lower-order approximation of the true Hamiltonian, and use backward error analysis to post-process the learned Hamiltonian to obtain a higher-order approximation of the true Hamiltonian. 
For the implicit midpoint method \eqref{eq:im_itegrator} that we use in our analysis, the modified Hamiltonian is given by
\begin{equation}\label{eq:modified_Hamiltonian_IM}
    \begin{aligned}
        \tilde{\Hc} = \Hc - \frac{h^2}{24} \nabla^2\Hc(J^{-1}\nabla \Hc, J^{-1}\nabla \Hc) + \mathcal{O}(h^4).
    \end{aligned}
\end{equation}
The expression $\nabla^2\Hc (J\nabla \Hc, J\nabla \Hc)$ represents a bilinear form that is explicitly given by
\begin{align*}
    \nabla^2\Hc (J^{-1}\nabla \Hc, J^{-1}\nabla \Hc) = (J^{-1}\nabla \Hc)^\top\nabla^2\Hc (J^{-1}\nabla \Hc).
\end{align*}
More about the modified Hamiltonian will be presented in the numerical section.

\subsection{Constructing symplectic integrators}\label{ss:prk}
\medskip
\paragraph{Partitioned Runge--Kutta methods\cite{jay1996symplectic}}
Partitioned Runge--Kutta (PRK) methods are a class of numerical integrators that are particularly well-suited for systems where the state can be naturally split into multiple components, such as position and momentum in Hamiltonian systems as given by: 
\begin{equation}\label{eq:prk_dyn}
\dfrac{dq}{dt}  = \frac{\partial\Hc(q,p,t)}{\partial p}, \quad \dfrac{dp}{dt}  = -\frac{\partial\Hc(q,p,t)}{\partial q}.
\end{equation}
Now generally, when we are solving for systems, we can combine these equations in a single vector of state $y(t)$ and iteratively solve for $y_n$ using RK methods, however, for Hamiltonian systems these variables play different roles and may evolve at different rates hence these methods treat these two set of variables differently. Now, equation \eqref{eq:prk_dyn} can be integrated using a partitioned Runge--Kutta scheme:
\begin{align*}
q_{n+1} &= q_n + h_n \sum_{i=1}^{s} b_i k_{n,i},\\ \qquad p_{n+1} &= p_n + h_n \sum_{i=1}^s B_i l_{n,i},
\end{align*}
where
\begin{align*}
k_{n,i} = f(Q_{n,i}, P_{n, i}, t_n + c_ih_n),
\\ \\
\qquad l_{n,i} = g(Q_{n,i}, P_{n,i}, t_n + C_ih_n).
\end{align*}
which are evaluated at the internal stages,
\begin{equation}
\begin{aligned}
Q_{n,i} = q_n + h_n \sum_{j=1}^{s} a_{ij}k_{n,j}, \qquad P_{n,i} = p_n + h_n \sum_{j=1}^s A_{ij} l_{nj}. 
\end{aligned}
\end{equation}
A partitioned Runge--Kutta scheme is symplectic if the following conditions hold:
\begin{equation}\label{eq:symp_condition}
\begin{aligned}
c_i = C_i, 
b_i = B_i, \qquad i &= 1, ..., s; \\ 
b_i A_{ij} + B_j a_{ji} - b_i B_j = 0, \quad i,j&=1,...,s. \\
\end{aligned}
\end{equation}
All the symplectic integrators discussed above, including the one used in our subsequent analysis, are examples of symplectic PRK schemes.

\paragraph{Hamiltonian variational integrators \cite{leok2011discrete}}
Discrete variational analysis is another way to construct symplectic integrators. These apply a discrete variational principle, which automatically guarantees symplecticity. In fact, discrete Lagrangian mechanics is expressed in terms of a discrete Lagrangian $L(q_k, q_{k+1})$ that is a Type I generating function and can be used to generate symplectic maps. However, if the Hamiltonian is not hyperregular and the Legendre transformation is not invertible, as is the case for $\Hc = qp$, it is not possible to transform the Hamiltonian into an equivalent Lagrangian and apply the discrete Lagrangian construction. Instead, it was showed in \cite{leok2011discrete} that a discrete Hamiltonian $H^+_d(q_k, p_{k+1})$ that is a Type II generating function can be constructed directly from the continuous Hamiltonian, thereby avoiding the need to evoke the Lagrangian perspective. As with the Lagrangian theory, the discrete Hamiltonian perspective extremizes a discrete action sum, yielding a Hamiltonian variational integrator that is automatically symplectic.

\section{Fully symplectic Hamiltonian Neural Network}
\label{sec:method}
\medskip
Now, we are at the point where we can apply the insights from symplectic integration theory to the data-driven setting.
Consider the problem of learning a Hamiltonian 
$\Hc$ from data, with noisy observations $(\tilde{\bq}, \tilde{\bp}) \in \RR^{2d}$. At this point let us set up a few assumptions
\begin{itemize}
    \item We have a set of noisy trajectories $N_{train}$ in the training set and $N_{val}$ in the validation set.
    \item Our integrator is represented by a flow map $\Phi$ such that $y_T = \Phi(y, f(y), 0, T, h)$, where $y$ is the initial state, $f(y)$, the generator of dynamics, $T$ is the final time and $h$ is stepsize of the integrator.
    \item The Hamiltonian Neural Network is represented by $\Hc_{\theta}(\bq, \bp)$ where $\theta\in \RR^p$, $(\bq, \bp)\in \RR^{2d}$ is a combined vector of phase space coordinates and are also the inputs to our neural network. 
\end{itemize}

Now, the problem boils down to learning the functional form of Hamiltonian $\Hc_{\theta^*}(\bq, \bp)$ where inference in the network, for given trained $\boldsymbol{\theta}^*$, amounts to computing the value of Hamiltonian function $\Hc_{\theta^*}(\bq, \bp)$ over phase space at a point $(\bq, \bp) \in \RR^{2d}$. We start at a set of initial phase space coordinates $(\bq, \bp)$ and simulate the trajectory forward using a symplectic integrator to compute up to final time $T$ to get $\{\bq_T, \bp_T\}$.  At this stage, we construct a loss function 
\begin{equation}\label{eq:loss_function}
    \begin{aligned}
        Loss = \frac{1}{n}\Big(\sum_{i=1}^N\|\bq^i_T - \tilde{\bq}^i_T\|^2 + \|\bp^i_T - \tilde{\bp}^i_T\|^2\Big).
    \end{aligned}
\end{equation}
Here, $i$ represents a data sample/trajectory in phase space and $n$ is the batch size since we compute a gradient estimate over a minbatch. The vector $(\tilde{\bq},\tilde{\bp})$ represents the noisy ground truth values.
\begin{figure*}[!ht]
    \centering
    \includegraphics[width=\textwidth]{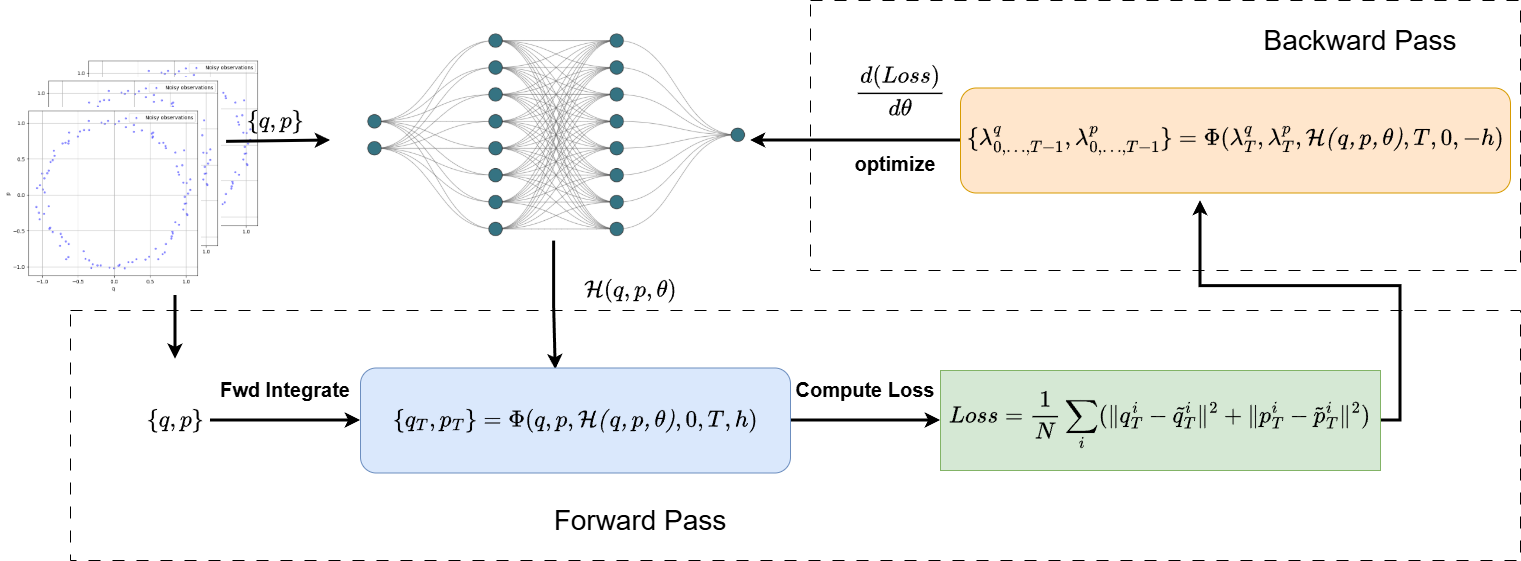}
    \caption{The schematic of the Hamiltonian Identification framework, where the network represents a parametrized Hamiltonian, the block in the blue below represents the ODE solver in the forward pass, and the block in the orange above represents the same ODE solver but for adjoint dynamics to get the loss gradients.}
    \label{fig:HNN_schematic}
\end{figure*}

\begin{algorithm}[H]
\caption{Hamiltonian identification and optimization}\label{alg:hamiltonian_identification}
\begin{algorithmic}[1]
\State \textbf{Input:} Initial $\boldsymbol{\theta}_{0}$, Train data $\{(\tilde{\bq}^i,\tilde{\bp}^i)\}^{i=0}_{N_{train}},\text{Val data} \{(\tilde{\bq}^i,\tilde{\bp}^i)\}^{i=0}_{N_{val}}$, learning rate $\eta$, \texttt{max\_epochs}
\State \textbf{Output:} Optimized parameter $\boldsymbol{\theta}^{*}$

\For{$k = 0$ to \texttt{max\_epochs}}
  \State \textbf{Forward pass}
  \State Define $\Hc_{\theta}(\bq,\bp)$ and get $\nabla_{\bq}\Hc,\nabla_{\bp}\Hc$ using autodiff
  \State $(\bq,\bp) \gets (\tilde{\bq},\tilde{\bp})$ \Comment{initialize state with noisy i/p data}
  \For{$i = 0$ to $N-1$}
    \State $(\bq_{T},\bp_{T}) \gets \Phi\!\big(\bq,\bp;\Hc_{\theta_k}(\bq,\bp), 0, T, h\big)$ \Comment{implicit map \eqref{eq:im_int}, see Alg.~\ref{alg:pc-fpi}}
  \EndFor
  \State $Loss(\boldsymbol{\theta}_{(k)}) \gets \frac{1}{n}\sum_{i=1}^{n}\!\left(\|\bq^i_T-\tilde{\bq}^i_T\|_2^2+\|\bp^i_T-\tilde{\bp}^i_T\|_2^2\right)$ \Comment{Compute loss}

  \State \textbf{Adjoint backward pass}
  \State $\boldsymbol{\lambda}^q_N \gets -\partial Loss / \partial \bq_N$;\quad
         $\boldsymbol{\lambda}^p_N \gets -\partial Loss / \partial \bp_N$ \Comment{Adjoint terminal states, cf.~\eqref{eq:adj_terminal_condns}}
  \For{$i = N$ to $1$}
    \State $(\boldsymbol{\lambda}^q_{\{T-1,\dots,0\}},\boldsymbol{\lambda}^p_{\{T-1,\dots,0\}})
      \gets \Phi\!\big(\boldsymbol{\lambda}^q_T,\boldsymbol{\lambda}^p_T;\Hc_{\theta_k}(\bq,\bp), T, 0, -h\big)$ \Comment{adjoint implicit map}
  \EndFor
  \State $g_k \gets  d (Loss) / d \boldsymbol{\theta}_{k}$ \Comment{Loss gradient \eqref{eq:lossgrad}}

  \State \textbf{Parameter update}
  \State $\boldsymbol{\theta}_{k+1} \gets \boldsymbol{\theta}_{k} - \eta\, g_k$
\EndFor
\State \textbf{Return:} $\boldsymbol{\theta}^{*}, \Hc_{\theta^*}(\bq, \bp)$
\end{algorithmic}
\end{algorithm}

\footnotetext{The quantities like $\lambda\frac{\partial^2 \Hc}{\partial p\partial q}$ and $\lambda\frac{\partial^2 \Hc}{\partial q\partial p}$ are computed via vector-Jacobian products in reverse-mode automatic differentiation, avoiding explicit Hessian construction.}

\subsection{Numerical forward pass with implicit-midpoint method}\label{ss:int_impl}
\medskip
In the forward pass, as represented by lower block in schematic \ref{fig:HNN_schematic}, we solve the Hamiltonian ODEs in equation \eqref{eq:hamilton_odes} to generate the forward trajectory up to final time $T$.

\begin{equation}\label{eq:hamilton_odes}
\begin{aligned}
\dfrac{dq_i}{dt}  = \frac{\partial \Hc}{\partial p_i}, \quad \dfrac{dp_i}{dt}  = -\frac{\partial \Hc}{\partial q_i}.
\end{aligned}
\end{equation}

As a choice of integrator, we choose the second-order implicit midpoint method with updates given by equation \eqref{eq:im_int}, 
\begin{equation}\label{eq:im_int}
\begin{aligned}
    q_{i+1} &= q_i + h\nabla_p\Hc_{\theta}\left(\frac{q_{i} + q_{i+1}}{2},\frac{p_{i} + p_{i+1}}{2}\right),\\
    p_{i+1} &= p_i - h\nabla_q\Hc_{\theta}\left(\frac{q_{i} + q_{i+1}}{2},\frac{p_{i} + p_{i+1}}{2}\right).
\end{aligned}
\end{equation}
The above equations are implicit in the unknowns and thus requires a fixed-point iteration scheme to solve. The details of the integrator with fixed-point iteration is discussed in Algorithm \ref{alg:pc-fpi}.

\begin{algorithm}[ht]
\caption{Predictor-Corrector with fixed-point iteration (PC + FPI)}
\label{alg:pc-fpi}
\begin{algorithmic}[1]
\Require Current state $\by \equiv (\bq, \bp)$, stepsize $h$, dynamics $\nabla_\by \Hc$, number of fixed-point iterations $\tau$
\State $\by' \gets \mathrm{RK2}(\by)$ \Comment{Predictor}
\For{$j = 1$ to $\tau$}
    \State $\by' \gets \by' + h \cdot \nabla_\by \Hc\left(\frac{\by + \by'}{2}\right)$ \Comment{Implicit Midpoint Update}
    \State $\by \gets \by'$
\EndFor
\State \Return $\by'$
\end{algorithmic}
\end{algorithm}

Starting from an initial point $\by \equiv (\bq, \bp)$, a preliminary estimate for the next step $\by'$ is first obtained using an explicit predictor, followed by refinement through repeated application of the implicit midpoint update via fixed-point iteration. Alternatively, when access to the full trajectory $(\bq_i, \bp_i)$ is available, as is the case in training against noisy trajectories, one may optionally bypass the predictor step and directly use the noisy values as an initial guess.

In our implementation, the simulation starts from an initial point 
$\by \equiv (\bq, \bp)$. 
Algorithm~\ref{alg:pc-fpi} is first applied with an explicit update to obtain a 
predicted value $\by'$, which is then used as the initial guess for the fixed-point iteration. For simulating up to final time $T$, each solve runs for $\tau$ fixed-point iterations.

\subsection{Backward pass with gradient computation via the adjoint sensitivity method}\label{ss:adjoint_sens}
\medskip

To perform an optimization over parameters $\theta$ to obtain the final $\theta^*$, we need to compute the sensitivity of the loss function w.r.t the parameters. As discussed in the previous section, pairing the ODE solver with fixed-point iterations incurs multiple function evaluations per timestep. To compute gradients for optimization, the automatic differentiation engine must retain the entire sequence of intermediate operations, to backpropagate gradients through the solver and evaluate the sensitivity of the loss with respect to $\boldsymbol{\theta}$.
This standard approach, often implemented via reverse-mode automatic differentiation (backpropagation), is memory-intensive, as it requires storing all intermediate values of $(\bq_t, \bp_t)$ and internal solver states across the entire integration window. This memory overhead grows linearly with the number of timesteps and becomes prohibitive for long-horizon simulations or high-dimensional dynamical systems.

To overcome this bottleneck, one can instead formulate the gradient computation as a boundary-value problem via the adjoint sensitivity method. This method derives a backward-in-time differential equation for the adjoint variables (or co-states), which represent the sensitivity of the loss functional with respect to the trajectory. Crucially, the backward pass is decoupled from the original forward pass: the adjoint equations can be solved without retaining the entire forward trajectory. This reformulation allows memory-efficient gradient computation, particularly when the number of scalar outputs (loss terms) is much smaller than the number of parameters.


In this section, we merely give the form of adjoint equations that we solve to get the gradient of the loss function w.r.t network parameters. In the next section, we give a detailed discussion of this approach. 
Given a neural network parameter estimation model subject to Hamiltonian ODEs
\begin{mini}|l|
 {\theta}{\Jc(\boldsymbol{\theta}, q, p) = \frac{1}{n}\sum \limits_{i=0}^n \|q^i_T - \tilde{q}^i_T\|^2_2 + \|p^i_T - \tilde{p}^i_T \|^2_2}{}{}
\addConstraint{\dot{q}(t) - \frac{\partial\mathcal{H}_{\theta}(q, p)}{\partial p}}{=0, \hspace{1cm} q(0) = q_0}
\addConstraint{\dot{p}(t) + \frac{\partial\mathcal{H}_{\theta}(q, p)}{\partial q}}{=0, \hspace{1cm} p(0) = p_0},
\label{eq:opt_prob_1}
\end{mini}
where $(q_T, p_T)$ is the predicted value of states through the symplectic map $\Phi_h$ at final time $T$, $h$ is the stepsize, and $(\tilde{q}_T, \tilde{p}_T)$ is the true value of those variables. For simplicity, here we consider $q$ and $p$ as scalar functions. 
Now, considering $\Phi_h$ as our symplectic map \eqref{eq:im_int} and using a shorthand notation $y_t$ for $(q_t, p_t)$, we can write
\begin{align*}
       \by_t = \Phi_h(\by_t, \by_{t-1}).
\end{align*}
In the usual cases where a neural network is used as a function approximator, the gradients are computed using backpropagation which is a standard way to optimize the loss function by adjusting the network weights. In our case, the forward pass includes a symplectic ODE solver which means we have to backpropagate through this solver which will drastically increase the memory requirements, which can be understood if we consider our loss function
\begin{align*}
    \mathcal{J} = \frac{1}{n}\sum\limits_{i=1}^n (y_T - \tilde{y}_T)^2,
\end{align*}
where each $y_T$ is obtained via forward propagation through a symplectic ODE solver. It is clear that if we are solving for larger simulation times, the number of evaluations in the backward pass will scale with the number of timesteps as shown below:
\begin{align*}
    \frac{d\mathcal{J}}{d\boldsymbol{\theta}} =  \sum\limits_{j=1}^t\Big( \prod\limits_{k=j+1}^t\frac{\partial y_k}{\partial y_{k-1}}\Big)\frac{\partial y_j}{\partial \theta},
\end{align*}
where each $y_t$ is a function of $y_{t}, y_{t-1}, y_{t-2},\dots,y_{1}$.  In contrast to this, if we use adjoint sensitivity analysis, we first have to derive the adjoint equations which is a system of ODEs in the adjoint variable $\lambda(t)$ which is a time dependent version of Lagrange multiplier which get introduced while solving equation \eqref{eq:opt_prob_1} using variational calculus. 
Given Hamiltonian ODEs in matrix notation
\begin{equation}\label{eq:Hamiltonian_system}
    \begin{aligned}
        \frac{d}{dt}
        \begin{pmatrix}
            q(t) \\
            p(t) 
        \end{pmatrix} =
        \begin{pmatrix}
            \nabla_p \Hc(q, p, \boldsymbol{\theta}) \\
            -\nabla_q \Hc(q, p, \boldsymbol{\theta})
        \end{pmatrix} := f(q, p, \theta),
    \end{aligned}
\end{equation}
the system of adjoint ODEs is then given as:
\begin{equation}\label{eq:adj_eqns}
\begin{aligned}
\frac{d}{dt} \begin{pmatrix}
\lambda^q \\
\lambda^p \\
\end{pmatrix}
&= -\begin{pmatrix}
\nabla_{qp} \Hc &  -\nabla_{qq} \Hc \\
\nabla_{pp} \Hc & -\nabla_{pq} \Hc
\end{pmatrix}
\begin{pmatrix}
\lambda^q \\
\lambda^p \\
\end{pmatrix}.
\end{aligned}
\end{equation}

Solving these equations backward in time (i.e. t=T to t=0) subject to terminal conditions:
\begin{equation}\label{eq:adj_terminal_condns}
\begin{aligned}
    \lambda^q(T) = -\frac{\partial \Jc}{\partial q}_{(t=T)}\text{ and } \quad \lambda^p(T) = -\frac{\partial \Jc}{\partial p}_{(t=T)},
\end{aligned}
\end{equation}
gives the adjoint state which is in turn used to calculate the loss gradients by computing the following integral:
\begin{equation}\label{eq:lossgrad}
     \frac{d\Jc}{d\theta} = \frac{\partial \Jc}{\partial \theta}-\int_0^T \lambda^T \frac{\partial f(q, p, \theta)}{\partial \boldsymbol{\theta}} dt.
\end{equation}
In \eqref{eq:lossgrad}, $\frac{d}{d\theta}$ denotes the total derivative, which accounts for the dependence of the trajectory $(q(t), p(t))$ on $\boldsymbol{\theta}$, whereas $\frac{\partial \Jc}{\partial \theta}$ denotes the explicit partial derivative of $\Jc$ with the trajectory held fixed. For the terminal loss \eqref{eq:loss_function}, $\Jc$ depends on $\boldsymbol{\theta}$ only through the trajectory, so $\frac{\partial \Jc}{\partial \theta}=0$ and the gradient reduces to $\frac{d\Jc}{d\theta} = -\int_0^T \lambda^T \frac{\partial f}{\partial \boldsymbol{\theta}}\, dt$; we retain the general form \eqref{eq:lossgrad} for consistency with the adjoint literature \cite{cao2003adjoint}.
Introducing the adjoint sensitivity method for gradient calculation will amount to solving the adjoint equations backward in time which is a constant memory task where we only need to store the current variable and its partial derivatives in the memory at any particular instant. The complete derivation of HNN adjoint state and gradients is provided in the subsequent section.

Notice that to get the adjoint state, we have to solve equation \eqref{eq:adj_eqns} subject to the terminal conditions \eqref{eq:adj_terminal_condns} $t=T$ to $t=0$ and store the results $\lambda(t)$ in memory to later compute the integral. Now compare it to backpropagation, where the memory requirements scale with simulation length $T$, where we have to store all intermediate sensitivities $\frac{\partial \by_i}{\partial\by_j}$, $0\leq j\leq i$, but here, we only need to solve the two equations for each pair of states, hence a constant memory operation\footnote{An important point to note is that the continuous adjoint computes gradients of the underlying ODE, whereas backpropagation corresponds to the discrete adjoint (cotangent lift) of the numerical integrator; these coincide only in the limit of vanishing stepsize or when the adjoint discretization is the symplectic (cotangent) lift of the forward scheme.}.  
\subsection{Derivation of adjoint sensitivity for Hamiltonian system from variational perspective}\label{ss:adjoint_proof}
\medskip
Here, we will present a proof of adjoint sensitivity rooted in variational analysis, which follows a similar line of reasoning as \cite{cao2003adjoint}. For that consider a continuous version of the problem that we are solving:
\begin{mini}|l|
 {y}{\Jc(\theta, y) = \int \limits_0^T f(\theta, y, t)dt}{}{}
\addConstraint{\dot{y} - f(y, \theta)}{=0}
\addConstraint{y(0)}{=y_0}.
\label{eq:opt_prob_adj}
\end{mini}
where $\theta\in \RR^p$ is a vector of unknown parameters, $y\in \RR^{2d}$ is a vector of state variables  and $h:\RR^{2d}\rightarrow\RR^{2d}$ is the forward dynamics.

\medskip


In our case, we use a forward map $\Phi_h$ to generate the flow $y_1, y_2\dots y_T$ and evaluate the loss/mismatch at the terminal point $y_T$.

The loss over a batch of trajectories can be written as:
\begin{align*}
Loss = \frac{1}{N} \sum_{i=1}^N \|y_i(T) - \tilde{y}_i(T)\|^2.
\end{align*}
In the continuous limit, this can be written as
\begin{align*}
\mathcal{J}(y) = (y(T) - \tilde{y}(T))^2.
\end{align*}

Writing the Lagrangian for this, we get

\begin{equation}\label{eq:lagrangian}
\begin{aligned}
\mathcal{L}
= \Jc(y(T))
+ \int_0^T \lambda^T(\dot{y} - f(y,\theta)) dt
+ \mu^T(y(0) - y_0).
\end{aligned}
\end{equation}
Taking variation w.r.t $y$ of the Lagrangian and  following a similar procedure as above after integration by parts gives us

\begin{align*}
\delta \Lc
&= \frac{\partial \Jc(y(T))}{\partial y}\delta y(T)
- \int_0^T\Big( \dot{\lambda}
+  \Big(\frac{\partial f}{\partial y}\Big)^T\lambda \Big)^T\delta y \, dt + \Big[\lambda^T\delta y\Big]_0^T.
\end{align*}
Collecting terms, and using the fact that the initial condition $y(0)=y_0$ is prescribed, so that $\delta y(0)=0$, we get
\begin{align*}
\delta \Lc
&= \Big(\frac{\partial \Jc(y(T))}{\partial y} + \lambda(T)^T \Big)\delta y(T)
- \int_0^T\Big( \dot{\lambda}
+  \Big(\frac{\partial f}{\partial y}\Big)^T\lambda \Big)^T\delta y \, dt.
\end{align*}

Now, by the fundamental theorem of the calculus of variations, the coefficients of $\delta y$ must vanish, which yields the adjoint equation
\begin{equation}\label{eq:adjoint_dynamics_2}
\begin{aligned}
    \dot{\lambda}(t) = -\left(\frac{\partial f}{\partial y}\right)^T\lambda(t), \qquad \text{s.t.} \quad \lambda(T) =- \frac{\partial \Jc(T)}{\partial y(T)}.
\end{aligned}
\end{equation}

\paragraph{Gradient w.r.t parameters:}

During the optimization, we require that the forward ODE and the boundary value constraint is always satisfied, formally:
\begin{align*}
\dot{y} - f(y,\theta) = 0,
\quad y(0) - y_0 = 0.
\end{align*}
\textbf{Note:} Numerically, the continuous ODE cannot be enforced exactly; instead, the discrete update equations of the symplectic integrator \eqref{eq:im_int} are satisfied exactly at every timestep (up to the tolerance of the fixed-point iteration), so the discrete trajectory satisfies the discrete constraints by construction, and along it the Lagrangian coincides with the loss. Symplecticity plays a complementary role: for a symplectic partitioned Runge--Kutta method, the discrete adjoint (cotangent lift) of the forward scheme coincides with the same scheme applied to the continuous adjoint system \eqref{eq:adjoint_dynamics_2}, so discretizing the adjoint equation with the same integrator yields the exact gradient of the discrete loss, in agreement with backpropagation.\\
Thus, the gradient of the loss w.r.t. the parameters is just the gradient of the Lagrangian, as the other two terms vanish identically along solutions of the forward system,
\begin{align*}
\frac{d \Jc}{d\theta} = \frac{d\mathcal{L}}{d\theta}.
\end{align*}
Differentiating the Lagrangian again in \eqref{eq:lagrangian} w.r.t. $\theta$ (network parameters):

\begin{align*}
\frac{d\mathcal{L}}{d\theta}
&= \frac{\partial \Jc}{\partial \theta}
+ \frac{\partial \Jc}{\partial y} \frac{\partial y}{\partial \theta}\bigg|_{t=T}
+\int_0^T
 \lambda^T\left(
\frac{\partial \dot{y}}{\partial \theta}
- \frac{\partial f}{\partial y} \frac{\partial y}{\partial \theta}
- \frac{\partial f}{\partial \theta}
\right)
 dt.
\end{align*}
Note that the second term does not vanish on its own; it will instead be cancelled by a boundary term below, once the terminal condition on $\lambda$ is imposed. Using integration by parts on $\int_0^T \lambda^T\frac{\partial \dot{y}}{\partial \theta}\,dt$, we obtain
\begin{align*}
\frac{d\mathcal{L}}{d\theta}
= \frac{\partial \Jc}{\partial \theta}
+ \frac{\partial \Jc}{\partial y} \frac{\partial y}{\partial \theta}\bigg|_{t=T}
+ \int_0^T \Bigg[
- \lambda^T \frac{\partial f}{\partial \theta}
 - \left( \dot{\lambda} + \Big(\frac{\partial f}{\partial y}\Big)^T\lambda\right)^T\frac{\partial y}{\partial \theta}\Bigg] dt + \Big[\lambda^T\frac{\partial y}{\partial \theta} \Big]^T_0.
\end{align*}
The coefficient of $\frac{\partial y}{\partial \theta}$ inside the integral is exactly the adjoint equation \eqref{eq:adjoint_dynamics_2}, so when restricted to solutions $\lambda(t)$ of the adjoint equation, it vanishes identically. At the boundaries, the initial condition $y(0)=y_0$ does not depend on $\theta$, so $\frac{\partial y(0)}{\partial \theta}=0$, while imposing the terminal condition $\lambda(T) = -\big(\frac{\partial \Jc}{\partial y(T)}\big)^T$ from \eqref{eq:adjoint_dynamics_2} makes the boundary term at $t=T$ cancel the term $\frac{\partial \Jc}{\partial y}\frac{\partial y}{\partial \theta}\big|_{t=T}$. These choices spare us from having to compute and store the sensitivities $\frac{\partial y(t)}{\partial \theta}$ of each intermediate state, which is a memory-intensive task.

So, by solving $\lambda(t)$ backward in time from the terminal data $\lambda(T)=-\big(\frac{\partial \Jc}{\partial y(T)}\big)^T$ along the adjoint equation, we obtain the following expression,

\begin{align*}
    \frac{d\mathcal{L}}{d \theta} = \frac{\partial \Jc}{\partial \theta} - \int_0^T\lambda^T\frac{\partial f}{\partial \theta}dt,
\end{align*}
which is also the gradient of the loss function w.r.t. parameters $\theta$ as reasoned above. Therefore,
\begin{equation}
\begin{aligned}
\frac{d\Jc}{d\theta}
= \frac{\partial \Jc}{\partial \theta}- \int_0^T \lambda^T \frac{\partial f}{\partial \theta} dt.
\end{aligned}
\label{grad_case2}
\end{equation}

\paragraph{Symplecticity analysis}

If we look at the form of the adjoint dynamics for this case 
\begin{equation}\label{eq:definitions}
\begin{aligned}
    y \equiv \begin{pmatrix}
        q \\ p
    \end{pmatrix}, \qquad f\equiv \begin{pmatrix}
        \nabla_p\Hc \\
        -\nabla_q \Hc
    \end{pmatrix}, \qquad \frac{\partial f}{\partial y} \equiv \begin{pmatrix}
        \nabla_{qp}\Hc & \nabla_{pp}\Hc\\
        -\nabla_{qq}\Hc &-\nabla_{pq}\Hc
    \end{pmatrix},
\end{aligned}
\end{equation}
using these definitions, we see that the adjoint equation can be written as :
\begin{align*}
\dot{\lambda} = - J^T \nabla^2 H \lambda \qquad \text{where }\quad J = \begin{pmatrix}
        0 & 1\\
        -1 & 0
    \end{pmatrix}.
\end{align*}
We see that the adjoint system has a symplectic structure.

\textbf{Remark:} It should be noted that, more generally, the combined state--adjoint system of any ODE is Hamiltonian. Given an ODE $\dot{y}=f(y)$, the state--adjoint dynamics are generated by the formal Hamiltonian $H(y,\lambda)=\lambda^T f(y)$, where $\lambda$ is the adjoint variable; the resulting flow on the cotangent bundle is the cotangent lift of the flow of $f$, and is therefore symplectic \cite{tran2024geometric}. In particular, the formal Hamiltonian can be derived from the Pontryagin maximum principle by considering the case of trivial controls and a purely terminal cost, in which case the running-cost contribution vanishes and the Pontryagin Hamiltonian reduces to $H(y,\lambda)=\lambda^T f(y)$.

\section{Numerical results}

\label{sec:numerics}
\medskip
In section \ref{ss:hamiltonian_systems}, we present the performance of our method on four representative Hamiltonian systems: the double-well potential, the coupled harmonic oscillator, the Hénon--Heiles system, and the 2-body Kepler problem. These systems are chosen to cover a range of problems, including separable and non-separable Hamiltonians. For training, the data is generated by sampling initial conditions from a uniform random distribution within a specified bounded domain. In contrast, for evaluation, the model is tested on test data sampled from 3 separate distributions different from training data in order to evaluate the ability of the learned dynamics to generalize. In subsequent section \ref{ss:modified_hamiltonian}, we compare our method with two baseline approaches in table \ref{tab:baseline_comparison}. All the computations were performed on the RCI SLURM cluster (2 × Intel Xeon Gold 6146, 46 cores $@$ 3.2 GHz, 384 GB RAM, 2 TB NVMe)\footnote{Our source code and details on how to reproduce the results is openly available at \texttt{https://github.com/choudharyharsh122/HNN}}.

\subsection{Implementation details }
\medskip
Below, we provide the main implementation details, including data generation and hyperparameters for our model.
\subsubsection{Data generation}
We generated 32,768 initial coordinates for the training set and 8,192 for the validation set. We used these initial conditions to generate ground truth trajectories using a fourth order symplectic integrator for 10 timesteps, where each trajectory serves as our ground truth data. 
Additive Gaussian noise $\mathcal{N}(0, 0.01)$ was introduced at each timestep to corrupt the trajectories with noise. The details on the distribution of data and sampling are plotted in the respective figures below.

\subsubsection{Model architecture details:}
\begin{figure}[h!]
    \centering
    \includegraphics[width=0.50\linewidth]{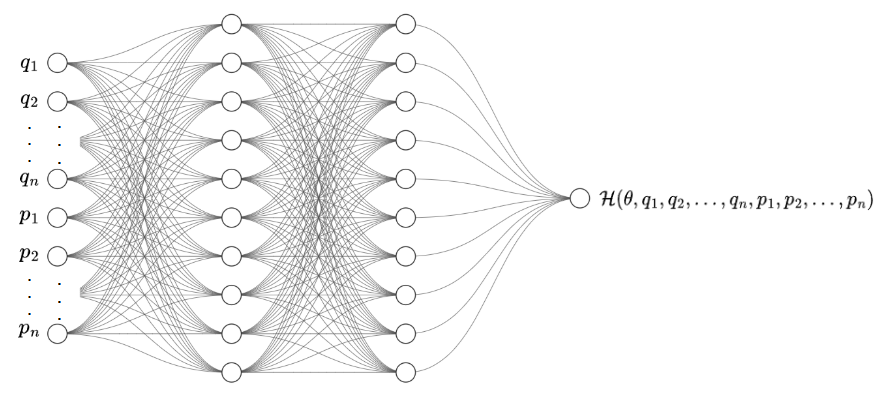}
    \caption{The model architecture of our neural network (the number of hidden layers and input dimension is configurable.)}
    \label{fig:hnn_model}
\end{figure}

\medskip
We parametrize the Hamiltonian ansatz with a neural network $\Hc_{\theta}(\bq, \bp)$ where $\bq$ and $\bp$ are one-dimensional vectors of canonical phase-space coordinates. Our neural network model has configurable input and hidden layers, which are passed as parameters. The network can learn a Hamiltonian of any number of particles in any number of dimensions as long as we have $dim(\bq) = dim(\bp)$ as shown in Figure \ref{fig:hnn_model}.

\subsubsection{Hyperparameter details}
\vspace{0.5em}
A minibatch size of 512 was used for both the training and validation datasets. The ansatz neural network has a configurable number of layers and dimensions passed by the user as command line arguments. For the first three experiments, we used a single hidden-layer NN model with dimension 32 and for the Kepler system, we use a hidden layer dimension 64. The input layer had dimension $2d$, which again is configurable, where $2d$ is the dimension of the phase space of system under study. Model parameters were optimized using the Adam optimizer with an initial learning rate of $10^{-3}$
 . A learning rate scheduler based on the ReduceLROnPlateau strategy was employed to adapt the learning rate during training in response to stagnation in validation loss. Hyperbolic tan is used as an activation function for all except for the final layer which is linear activation. For each batch, the starting point was sampled at a random timestep along the trajectory, allowing the model to observe a wide variety of initial conditions and improve generalization across the phase space. In the forward pass, we simulated trajectories for 2 timesteps and evaluated the loss according to \eqref{eq:final_objective}. Gradients were computed using the adjoint state method, employing the same integrator as used in the forward pass, which leverages the fact that a symplectic Runge--Kutta integrator is invariant under cotangent lift. The loss function is
\begin{equation}\label{eq:final_objective}
    Loss = \frac{1}{n}\sum\limits_{i=1}^{n} \|p^i_{T} - \tilde{p}^i_T\|_2^2 + \|q^i_T - \tilde{q}^i_T\|_2^2
\end{equation}
where the state variables $(q_t, p_t)$ are evolved using a symplectic integrator.

\subsubsection{Evaluation criteria}
\vspace{0.5em}
We aim to learn the functional form of the Hamiltonian \( \mathcal{H}(q, p) \) beyond simple trajectory matching. Unlike prior works such as Hamiltonian Neural Networks (HNN) and Neural ODEs, which primarily assess the learned models by comparing predicted and ground-truth trajectories, we explicitly evaluate the learned Hamiltonian function  $\mathcal{H}_{NN}$ across the broader phase space. To do so, we sample test points \( (q^i, p^i) \) from three distinct distributions over the phase space:

\begin{itemize}
    \item \textbf{Random Uniform:} Points are sampled independently from a uniform distribution over a fixed bounding box that encompasses the training trajectories. This probes generalization to randomly scattered unseen states.

    \item \textbf{Uniform Square Grid:} A structured grid of evenly spaced points is generated within the same bounding box. This enables a systematic and resolution-controlled evaluation of \( \mathcal{H}_{NN} \) over the phase space.

    \item \textbf{Gaussian:} Points are sampled from a multivariate Gaussian distribution centered around typical states observed during training.
\end{itemize}

The results for the Hamiltonian prediction error $\|\Hc_{true} - \Hc_{NN}\|_1$\footnote{$\Hc_{NN}$ here is the offset-corrected Hamiltonian as the neural network learns an approximation of $\Hc_{true}$ up to some global constant. To correct for that, we follow the same technique described in \cite{david2023symplectic}} over the test domain are plotted in Figures \ref{fig:h_error_dw}-\ref{fig:h_error_kepler} for the four Hamiltonian systems under evaluation, respectively.


\subsection{Hamiltonian systems and results}\label{ss:hamiltonian_systems}
\begin{figure}[!ht]
    \centering
    \begin{subfigure}{0.22\textwidth}
        \includegraphics[width=\textwidth]{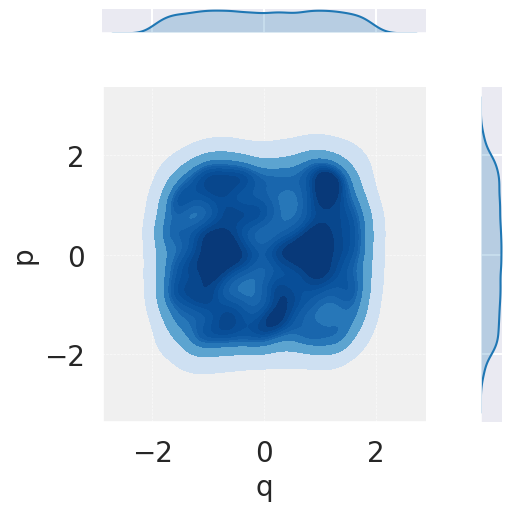}
        \caption{}
        \label{fig:ip_data_dw}
    \end{subfigure}
    \hfill
    \begin{subfigure}{0.22\textwidth}
        \includegraphics[width=\textwidth]{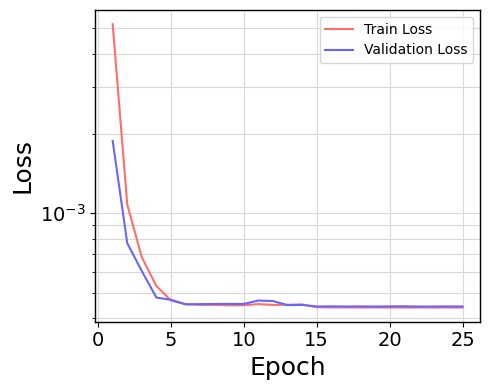}
        \caption{}
        \label{fig:loss_dw}
    \end{subfigure}\hfill
    \begin{subfigure}{0.22\textwidth}
        \includegraphics[width=\textwidth]{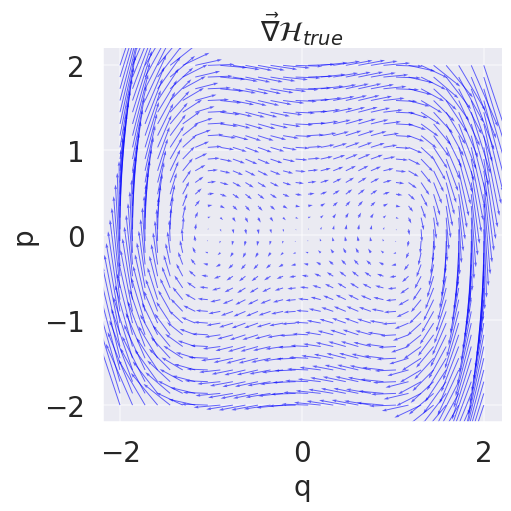}
        \caption{}
        \label{fig:true_dyn_dw}
    \end{subfigure}
    \hfill
    \begin{subfigure}{0.22\textwidth}
        \includegraphics[width=\textwidth]{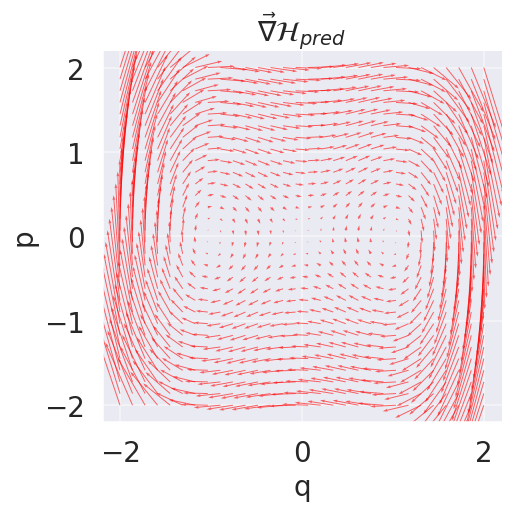}
        \caption{}
        \label{fig:pred_dyn_dw}
    \end{subfigure}
    \caption{Representative plots for (a) distribution of training data (b) training and validation loss (c) true dynamics (d) predicted dynamics for the double well potential.}
    \label{fig:double_well}
\end{figure}

\begin{figure}[!htp]
    \centering
    \begin{subfigure}{0.31\textwidth}
        \includegraphics[width=\textwidth]{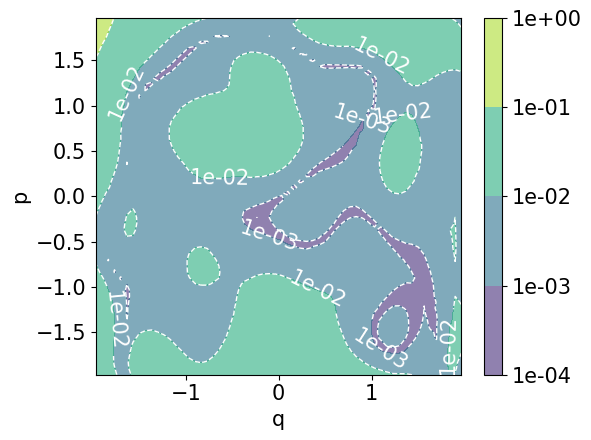}
        \caption{}
        \label{fig:dw_uniform}
    \end{subfigure}
    \hfill
    \begin{subfigure}{0.31\textwidth}
        \includegraphics[width=\textwidth]{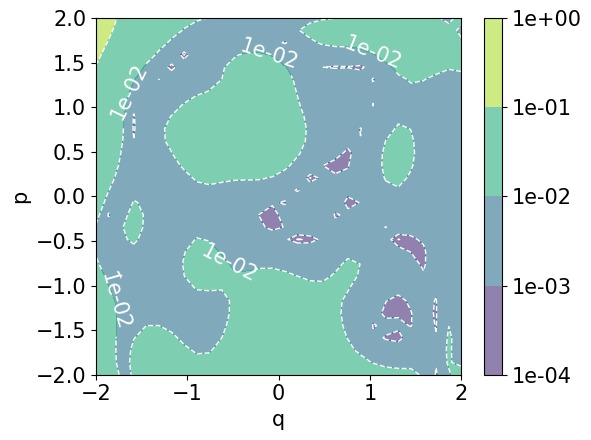}
        \caption{}
        \label{fig:dw_grid}
    \end{subfigure}
    \hfill
    \begin{subfigure}{0.31\textwidth}
        \includegraphics[width=\textwidth]{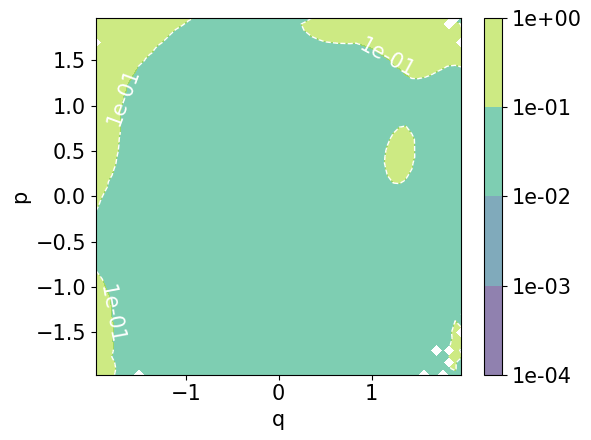}
        \caption{}
        \label{fig:dw_gaussian}
    \end{subfigure}
    \vspace{-4pt}
    \caption{Error contours for $\| \Hc_{true} - \Hc_{NN}\|_1$ for the double well system on test data drawn from 3 different distributions: (a) random uniform; (b) uniform square grid; (c) multivariate Gaussian \Nc(\textbf{0}, \textbf{$I_2$}).}
    \label{fig:h_error_dw}
\end{figure}

\medskip
\textbf{System 1: Double well potential} 
The particle in a double-well potential is a commonly studied system in classical and quantum mechanics where the system has 2 stable fixed points. In our case, we consider a symmetrical double potential well with Hamiltonian and governing equations given by
\begin{equation}\label{eq:state_dw}
\begin{aligned}
\Hc &= \frac{p^2}{2} + \frac{q^4}{4} - \frac{q^2}{2}, \\
 \dot{q}&= p,\quad
 \dot{p}= q - q^3.
 \end{aligned}
\end{equation}
We first generated the input data with a uniform distribution in the range $q \in[-2, 2]$ and $p\in[-2,2]$.  We then trained our neural network for 25 epochs, within which the validation loss converges sufficiently: the training and validation losses plateau at approximately $5\times10^{-4}$ within the first five epochs and remain essentially indistinguishable from each other thereafter, indicating that the model does not overfit the noisy trajectories.
The plots in Figure \ref{fig:double_well} show the input data distribution, training and validation loss, and the predicted dynamics for the double well system. The error in Hamiltonian prediction over phase space is shown in figure \ref{fig:h_error_dw}
\\

\begin{figure}[!ht]
    \centering
    \begin{subfigure}{0.22\textwidth}
        \includegraphics[width=\textwidth]{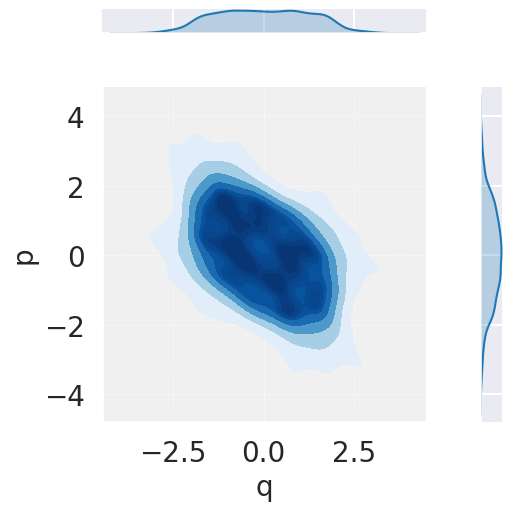}
        \caption{}
        \label{fig:ip_data_cho}
    \end{subfigure}
    \hfill
    \begin{subfigure}{0.22\textwidth}
        \includegraphics[width=\textwidth]{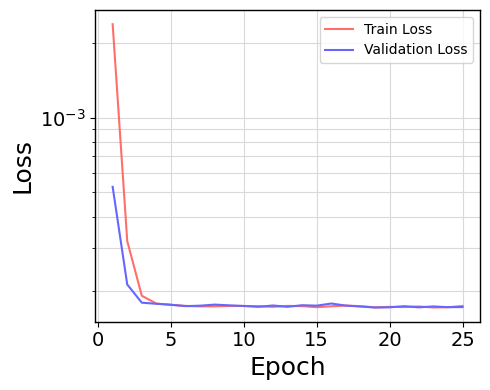}
        \caption{}
        \label{fig:loss_cho}
    \end{subfigure}\hfill
    \begin{subfigure}{0.22\textwidth}
        \includegraphics[width=\textwidth]{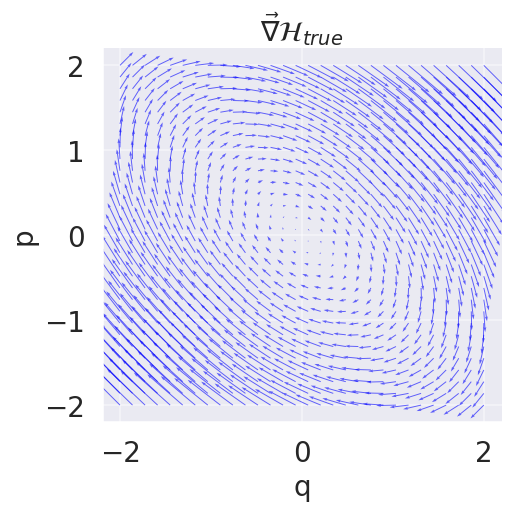}
        \caption{}
        \label{fig:true_dyn_co}
    \end{subfigure}
    \hfill
    \begin{subfigure}{0.22\textwidth}
        \includegraphics[width=\textwidth]{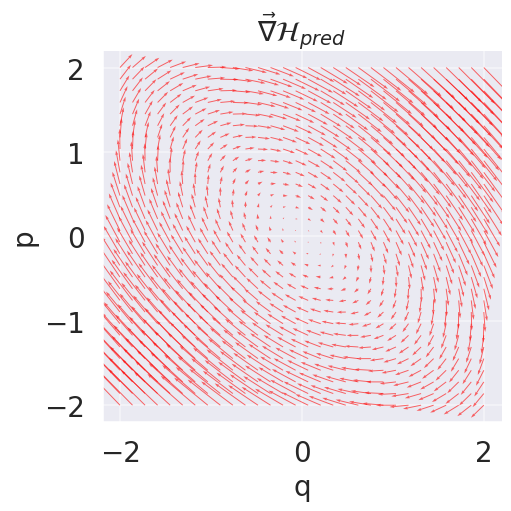}
        \caption{}
        \label{fig:pred_dyn_co}
    \end{subfigure}
    \caption{Representative plots for (a) distribution of training data (b) training and validation loss (c) true dynamics (d) predicted dynamics for the coupled harmonic oscillator.}
    \label{fig:coupled_ho}
\end{figure}

\begin{figure}[!htp]
    \centering
    \begin{subfigure}{0.31\textwidth}
        \includegraphics[width=\textwidth]{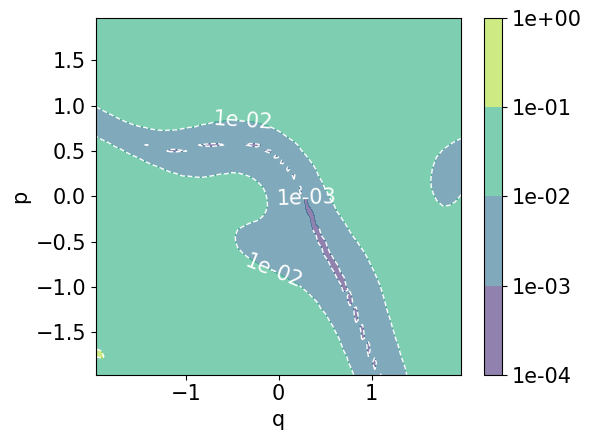}
        \caption{}
        \label{fig:cho_uniform}
    \end{subfigure}
    \hfill
    \begin{subfigure}{0.31\textwidth}
        \includegraphics[width=\textwidth]{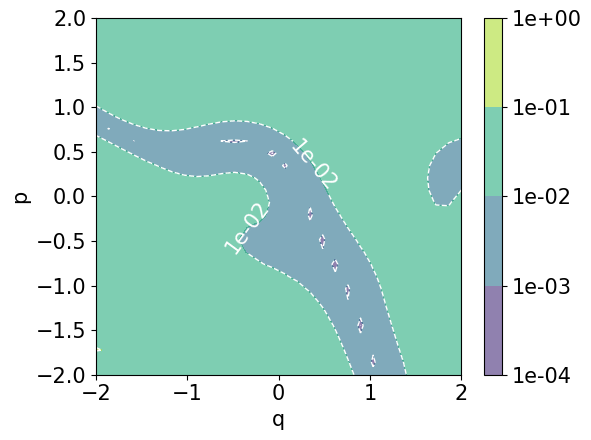}
        \caption{}
        \label{fig:cho_grid}
    \end{subfigure}
    \hfill
    \begin{subfigure}{0.31\textwidth}
        \includegraphics[width=\textwidth]{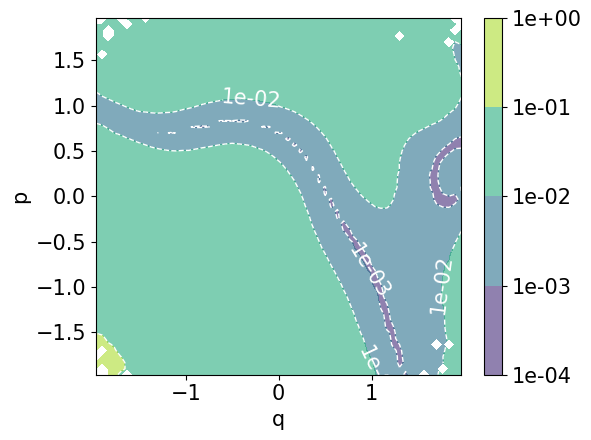}
        \caption{}
        \label{fig:cho_gaussian}
    \end{subfigure}
    \caption{Error contours for $\| \Hc_{true} - \Hc_{NN}\|_1$ for the coupled  oscillator system on test data drawn from 3 different distributions: (a) random uniform; (b) uniform square grid; (c) multivariate Gaussian \Nc(\textbf{0}, \textbf{$I_2$}).}
    \label{fig:h_error_cho}
\end{figure}

\noindent\textbf{System 2: Coupled Harmonic Oscillator}
\medskip
A coupled harmonic oscillator is a simple 1-D non-separable Hamiltonian system which consists of two simple harmonic oscillators which are coupled by an intermediary force or potential where $\alpha$ is the coupling constant. The Hamiltonian and governing dynamics given by
\begin{equation}\label{eq:state_coupled_ho}
\begin{aligned}
    \Hc =& \frac{p^2}{2} + \frac{q^2}{2} + \alpha pq,\\
    \quad  \dot{q}=& p + \alpha q, \quad  \dot{p}= -(q + \alpha p).
\end{aligned}
\end{equation}
The plots in Figure \ref{fig:coupled_ho} show the input data distribution, training and validation loss, and the predicted dynamics for the coupled oscillator system.
The system was trained for 25 epochs; we see a steep decrease in validation loss within a few initial epochs, indicating that the adjoint-based gradient estimates are sufficiently accurate to find right descent directions in objective. The training and validation losses plateau at approximately $2\times10^{-4}$ within four epochs and coincide thereafter. The error in Hamiltonian prediction over phase space is shown in Figure~\ref{fig:h_error_cho}.

\begin{figure}[!ht]
    \centering
    \begin{subfigure}{0.22\textwidth}
        \includegraphics[width=\textwidth]{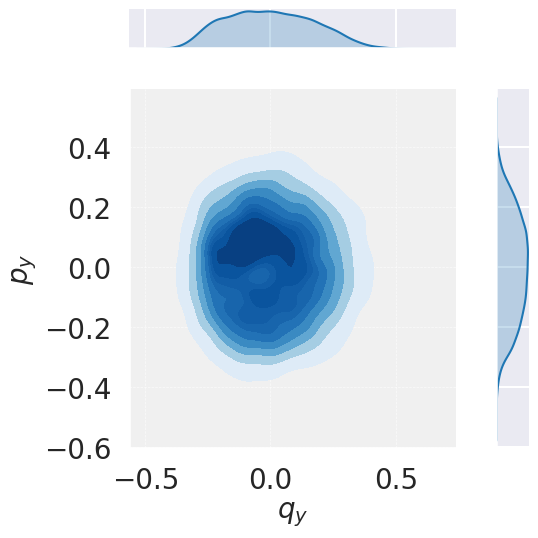}
        \caption{}
        \label{fig:ip_data_hh}
    \end{subfigure}
    \hfill
    \begin{subfigure}{0.22\textwidth}
        \includegraphics[width=\textwidth]{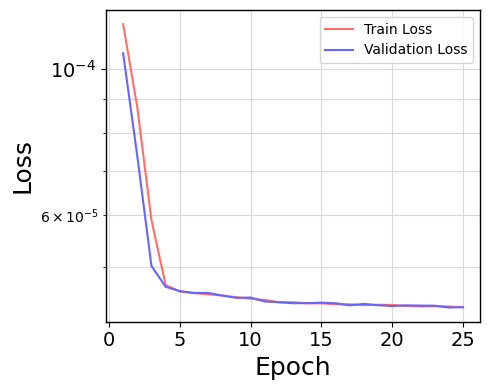}
        \caption{}
        \label{fig:loss_hh}
    \end{subfigure}\hfill
    \begin{subfigure}{0.22\textwidth}
        \includegraphics[width=\textwidth]{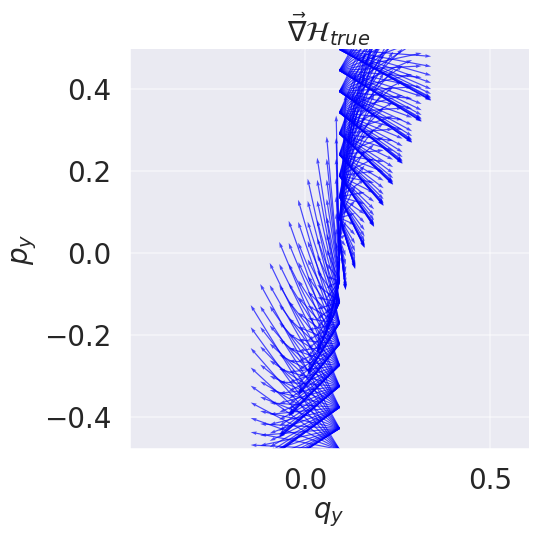}
        \caption{}
        \label{fig:true_dyn_hh}
    \end{subfigure}
    \hfill
    \begin{subfigure}{0.22\textwidth}
        \includegraphics[width=\textwidth]{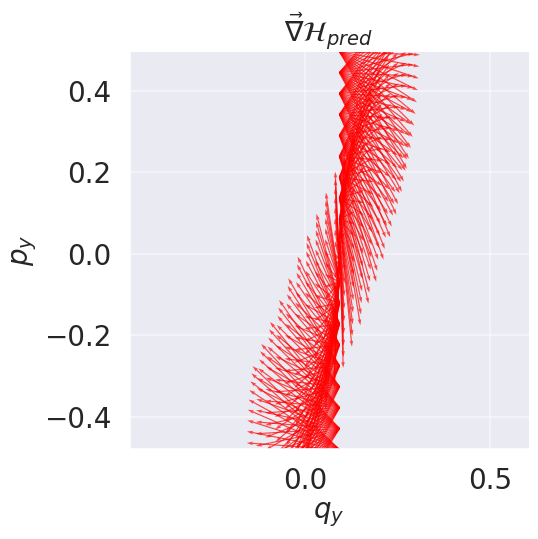}
        \caption{}
        \label{fig:pred_dyn_hh}
    \end{subfigure}
    \caption{Representative plots for (a) distribution of training data (b) training and validation loss (c) true dynamics (d) predicted dynamics for the H\'enon--Heiles system. Note that the $x$ and $y$ axes here represent projections of the $y$-coordinate of position and momentum for fixed $(p_x, q_x)$.}
    \label{fig:henon_heiles}
\end{figure}

\begin{figure}[!htp]
    \centering
    \begin{subfigure}{0.31\textwidth}
        \includegraphics[width=\textwidth]{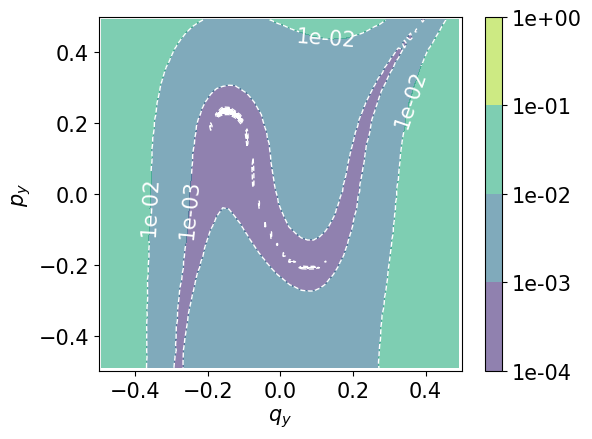}
        \caption{}
        \label{fig:hh_uniform}
    \end{subfigure}
    \hfill
    \begin{subfigure}{0.31\textwidth}
        \includegraphics[width=\textwidth]{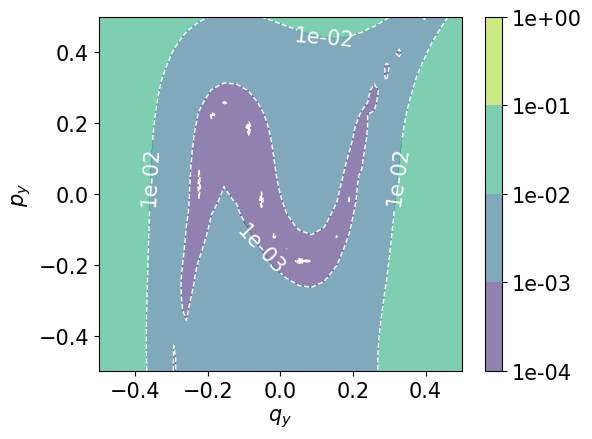}
        \caption{}
        \label{fig:hh_grid}
    \end{subfigure}
    \hfill
    \begin{subfigure}{0.31\textwidth}
        \includegraphics[width=\textwidth]{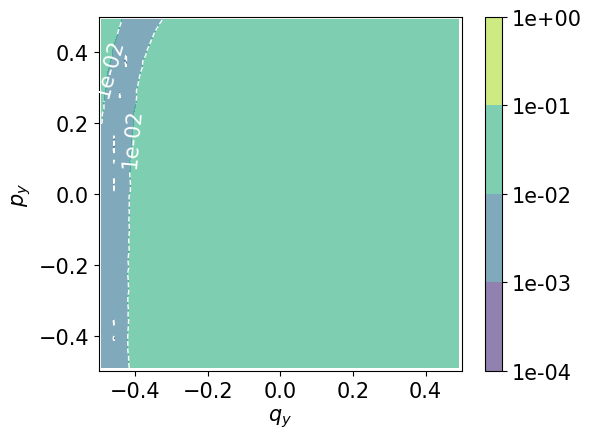}
        \caption{}
        \label{fig:hh_gaussian}
    \end{subfigure}
    \caption{Error contours for $\| \Hc_{true} - \Hc_{NN}\|_1$ for the H\'enon--Heiles system on test data drawn from 3 different distributions: (a) random uniform; (b) uniform square grid; (c) multivariate Gaussian \Nc(\textbf{0}, \textbf{$I_2$}). The plot is a 2D slice of the full 4D phase space, drawn by fixing the two coordinates $(q_x, p_x)$ and varying only $(q_y, p_y)$.}\label{fig:h_error_hh}
\end{figure}

\medskip
\noindent\textbf{System 3: Henon-Hieles Potential}
\medskip
We now explore higher-dimensional systems where chaos can emerge. Chaotic systems are deterministic yet unpredictable over long timescales due to exponential error growth, governed by the Lyapunov exponent. However, since these systems follow well-defined Hamiltonians, their dynamics can still be learned from limited observations.
A key example is the Hénon--Heiles (HH) system, a Hamiltonian system describing a star’s planar motion around a galactic center. While the system exhibits chaotic behavior, stable regions exist, aiding in learning its governing dynamics \cite{barrio2020distribution}. The Hamiltonian $\mathcal{H}$ and the corresponding equations of motion are given by 
\begin{equation}\label{eq:state_hh}
\begin{array}{ll}
\mathcal{H} = \frac{p_x^2 + p_y^2}{2} + \frac{q_x^2 + q_y^2}{2} + q_x^2q_y -\frac{q_y^3}{3},\\
 \dot{q_x} = p_x,\quad
 \dot{q_y} = p_y, \\
 \dot{p_x} = -q_x - 2q_xq_y,\quad
 \dot{p_y} = -q_y - q_x^2 + q_y^2.\\

 \end{array}
\end{equation}
The system was again trained for 25 epochs and the validation loss again converges quickly, plateauing at approximately $5\times10^{-5}$ within five epochs, with the training and validation curves nearly indistinguishable.
The plots in Figure \ref{fig:henon_heiles} show the input data distribution, training and validation loss, and the predicted dynamics for H\'enon--Heiles system. Figure \ref{fig:h_error_hh} shows the error in predicted Hamiltonian over a 2D phase space slice.

\begin{figure}
    \centering
    \includegraphics[width=0.35\linewidth]{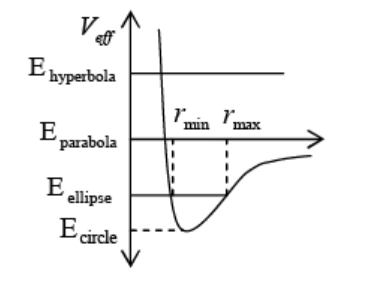}
    \caption{Various orbits in Kepler's potential}
    \label{fig:effective_potential}
\end{figure}

\begin{figure}[!htp]
    \centering
    \begin{subfigure}{0.22\textwidth}
        \includegraphics[width=\textwidth]{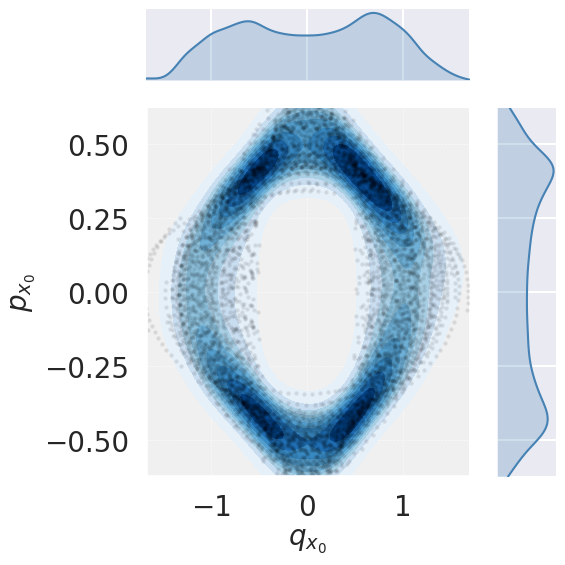}
        \caption{}
        \label{fig:ip_data_kepler}
    \end{subfigure}
    \hfill
    \begin{subfigure}{0.22\textwidth}
        \includegraphics[width=\textwidth]{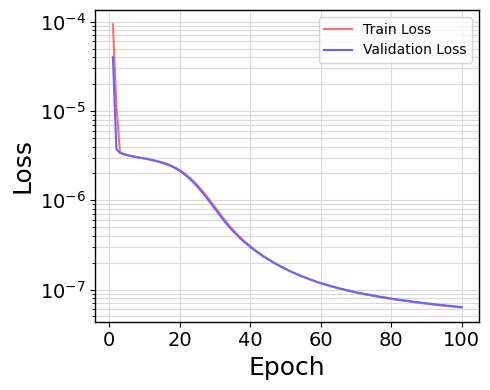}
        \caption{}
        \label{fig:loss_kepler}
    \end{subfigure}\hfill
    \begin{subfigure}{0.22\textwidth}
        \includegraphics[width=\textwidth]{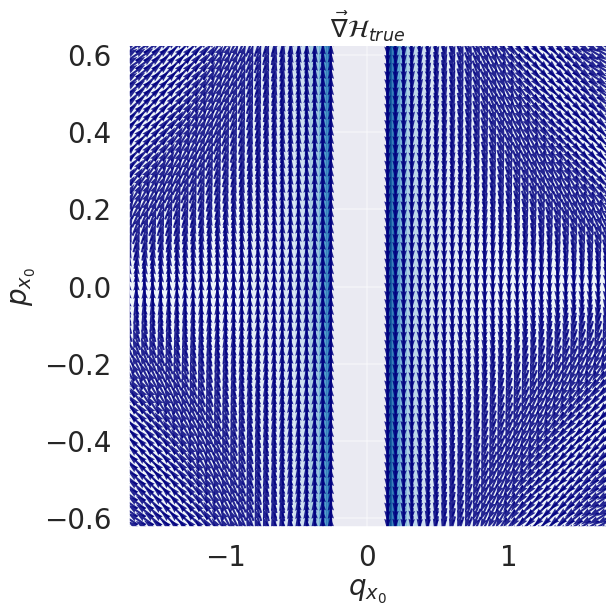}
        \caption{}
        \label{fig:true_dyn_kepler}
    \end{subfigure}
    \hfill
    \begin{subfigure}{0.22\textwidth}
        \includegraphics[width=\textwidth]{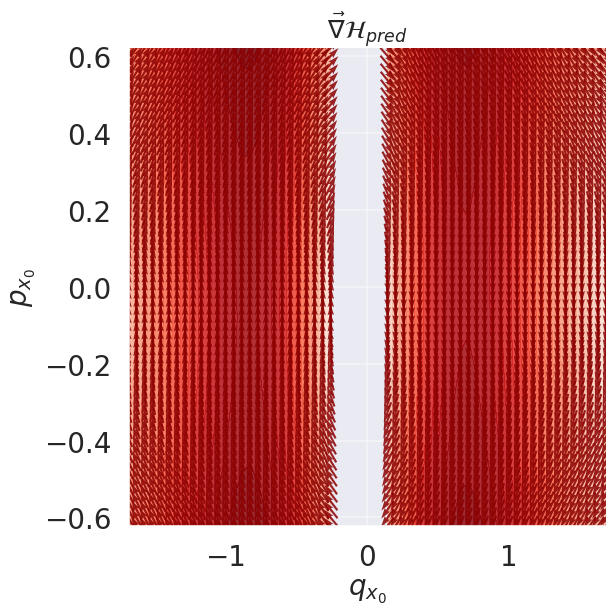}
        \caption{}
        \label{fig:pred_dyn_kepler}
    \end{subfigure}
    \caption{Representative plots for (a) distribution of training data (b) training and validation loss (c) true dynamics (d) predicted dynamics for the Kepler's potential. Note that the $x$ and $y$ axes here represent projections of the $x$-coordinate of position and momentum for particle 1.}
    \label{fig:kepler_problem}
\end{figure}

\begin{figure}[!ht]
    \centering
    \begin{subfigure}{0.31\textwidth}
        \includegraphics[width=\textwidth]{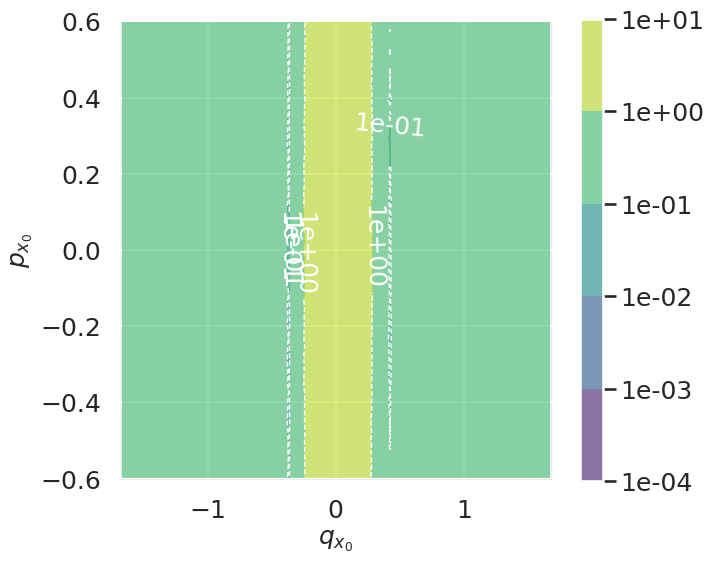}
        \caption{}
        \label{fig:kepler_uniform}
    \end{subfigure}
    \hfill
    \begin{subfigure}{0.31\textwidth}
        \includegraphics[width=\textwidth]{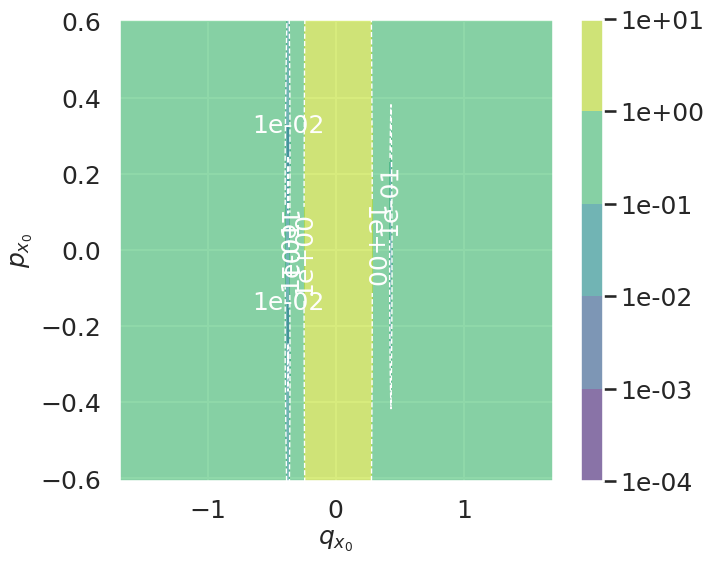}
        \caption{}
        \label{fig:kepler_grid}
    \end{subfigure}
    \hfill
    \begin{subfigure}{0.31\textwidth}
        \includegraphics[width=\textwidth]{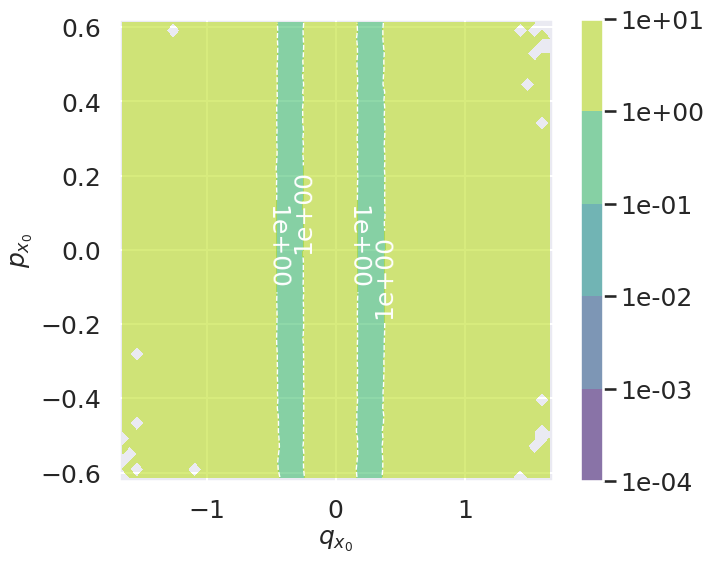}
        \caption{}
        \label{fig:kepler_gaussian}
    \end{subfigure}
    \caption{Error contours for $\| \Hc_{true} - \Hc_{NN}\|_1$ in Kepler's potential system on test data drawn from 3 different distributions: (a) random uniform; (b) uniform square grid; (c) multivariate Gaussian \Nc(\textbf{0}, \textbf{$I_2$}). Here, we fix all the other coordinates except for two coordinates $q_{x_0}, p_{x_0}$ which we sample from the three different distributions. Hence the plots represent a 2D slice of a 4D hyperplane in phase space.}
    \label{fig:h_error_kepler}
\end{figure}

\textbf{System 4: Kepler's potential}
\medskip
Kepler's potential is a spherically-symmetric central potential given by $V(r) = -\frac{k}{r}$, where $r$ is the radial distance between interacting particles. The problem has a singularity at $r=0$ but this is not a fundamental impediment to NN learning and symplectic integrators \cite{bolte2021conservative}. However, the trajectories exhibit finer scales and we have to significantly increase the sampling resolution to compensate for this. It represents the potential energy of a particle under gravitational or electrostatic attraction, resulting in elliptical, parabolic, or hyperbolic orbits. It describes Kepler's laws for planetary motion. These paths, described by Kepler's Laws, range from bound, closed elliptical orbits to unbounded, open parabolic or hyperbolic trajectories, depending on the system's total energy as shown in Figure~\ref{fig:effective_potential}. The total energy of a particle is given by
\begin{align*}
    E_{tot} &= \frac{1}{2}m\dot{r}^2 + V_{eff}, \qquad \text{where} \qquad V_{eff} = \frac{J^2}{2mr^2} -\frac{k}{r},
\end{align*}
where $J$ is the angular momentum and is a constant of motion. \\
We analyze the Hamiltonian of two interacting particles which,
in Cartesian coordinates of position and momenta, is given by
\begin{equation}\label{eq:kepler_2_body}
\begin{aligned}
    \Hc &= \frac{1}{2m_1}(p_{x_1}^2 + p_{y_1}^2) + \frac{1}{2m_2}(p_{x_2}^2 + p_{y_2}^2) - \frac{G m_1m_2}{\sqrt{(q_{x_1} - q_{x_2})^2 + (q_{y_1} - q_{y_2})^2}},\\
    \dot{q}_{x_1} &= \frac{p_{x_1}}{m_1},\quad \dot{q}_{y_1} = \frac{p_{y_1}}{m_1}, \quad \dot{q}_{x_2} = \frac{p_{x_2}}{m_2}, \quad \dot{q}_{y_2} = \frac{p_{y_2}}{m_2},\\
    \dot{p}_{x_1}&=-\frac{Gm_1m_2(q_{x_1} - q_{x_2})}{r^3_{12}}, \quad \dot{p}_{y_1}=-\frac{Gm_1m_2(q_{y_1} - q_{y_2})}{r^3_{12}},\\
    \dot{p}_{x_2}&=\frac{Gm_1m_2(q_{x_1} - q_{x_2})}{r^3_{12}}, \quad \dot{p}_{y_2}=\frac{Gm_1m_2(q_{y_1} - q_{y_2})}{r^3_{12}}.
\end{aligned}
\end{equation}
In the equations above, $q_{x_1}, p_{x_1}, q_{y_1}, p_{y_1}$ represent the phase space coordinates of particle 1 and similarly, $q_{x_2}, p_{x_2}, q_{y_2}, p_{y_2}$ represent the phase space coordinates for particle 2.\\
\medskip
\textbf{Data generation} Kepler's problem has a singularity at points where $\|q_i - q_j\|=0$ so to avoid sampling at the singularity, we sampled initial conditions such that $r\sim\Nc(\mu=4, \sigma=2.5)$ where r is the radial distance between the bodies and $\Nc(\mu, \sigma)$ represents a random normal distribution. The data was generated with a timestep of 0.01 sec with 100 total timesteps. After data generation, the trajectories were corrupted at each timestep with Gaussian noise sampled from $\Nc(\mu=0, \sigma=10^{-3})$. The neural network was trained for 100 epochs; the training and validation losses decrease by more than three orders of magnitude, to about $10^{-7}$, as seen in the training results presented in figure \ref{fig:kepler_problem}. Nevertheless, the accuracy of the learned Hamiltonian is worse as compared to other systems, as seen in Figure \ref{fig:h_error_kepler}: the error is largest for configurations with small interparticle separation, near the singularity of the potential, where the rapidly varying dynamics is undersampled by the trajectory data. The contrast between the small trajectory-matching loss and the larger Hamiltonian error indicates that, for this system, short trajectory segments in the sampled region only weakly constrain the global form of the Hamiltonian.


Figures \ref{fig:h_error_dw}--\ref{fig:h_error_kepler} allow a more quantitative assessment of the learned Hamiltonians. For the double well, the pointwise error is at or below $10^{-2}$ over most of the training domain and increases toward the boundary of the sampled region, where the trajectory data is sparse. For the coupled oscillator, the error dips below $10^{-3}$ in a band through the origin that is densely covered by training trajectories, and is on the order of $10^{-2}$ over the rest of the domain. For the H\'enon--Heiles system, the error over the plotted slice is below $10^{-3}$ in a large neighborhood of the origin, corresponding to the region of bounded, regular motion from which the training data is drawn, and grows toward $10^{-2}$ at the edges of the slice. For the Kepler problem, the error is on the order of $10^{-1}$ over most of the slice and is largest, of order one, in the band of configurations with small interparticle separation, consistent with the discussion of the singularity above. Comparing the three test distributions, the random uniform and uniform grid test sets yield comparable errors, while the Gaussian test set consistently produces the largest errors; this is expected, since the training initial conditions are sampled uniformly, so the Gaussian test set is furthest from the training distribution and, for the higher-dimensional systems, places part of its mass outside the training domain.

\subsection{Runtime and memory performance}\label{ss:mem_perform}
\begin{figure}[!ht]
    \centering
    \begin{subfigure}{0.45\textwidth}
        \includegraphics[width=\textwidth]{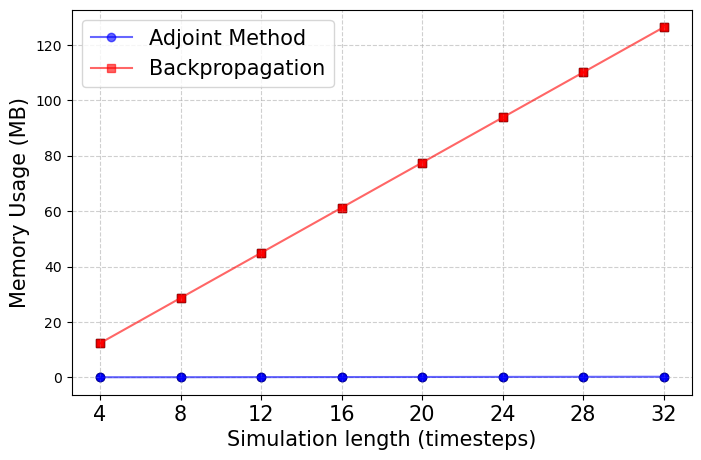}
        \caption{}
        \label{fig:subfig1_memory}
    \end{subfigure}
    \hfill
    \begin{subfigure}{0.45\textwidth}
        \includegraphics[width=\textwidth]{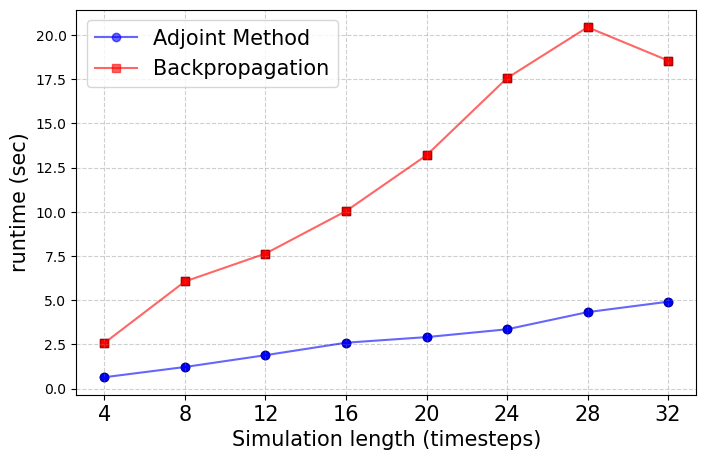}
        \caption{}
        \label{fig:subfig2_runtime}
    \end{subfigure}
    
    \caption{Comparison of (a) memory and (b) runtime profiles for adjoint and backdrop-based gradient evaluation. Each data point corresponds to the metrics evaluated for a single training iteration for a single batch of size 512 for the coupled harmonic oscillator system where the $x$-axis represents number of simulation timesteps of $h=0.01$ (Note that as the problem becomes larger, the runtime for adjoint state surpasses the backpropagation as it involves solving the terminal value problem in the backward pass.)}
    \label{fig:adj_vs_bprop}
\end{figure}

\medskip
In our study, we systematically analyzed the runtime and memory consumption of the adjoint method and backpropagation across increasing simulation lengths, ranging from 4 to 32 timesteps. The evaluation was conducted on a benchmark problem involving the training of a coupled harmonic oscillator system with a single batch of size 512. The results as shown in Figure \ref{fig:adj_vs_bprop} show a stark contrast in memory scalability between the two approaches: while the adjoint method maintains a constant memory footprint irrespective of the simulation length, the memory usage in backpropagation exhibits a linear growth pattern. This discrepancy arises due to the fundamental difference in how gradients are computed. Backpropagation explicitly stores intermediate states for every timestep, whereas the adjoint method reconstructs gradients via a reverse-time integration of the system dynamics, circumventing the need for extensive memory allocation. Interestingly, our runtime analysis also favors the adjoint method for smaller-scale problems, where it demonstrates superior computational efficiency, surpassing backpropagation in execution speed. 

\subsection{Sources of error in Hamiltonian learning}\label{ss:sources_of_error}
\medskip
The task of Hamiltonian learning using a symplectic integrator has three main sources of error 
\begin{itemize}
    \item \textbf{Approximation error}: Due to the limit of expressivity of our neural network model.
    \item \textbf{Learning error}: Due to incorrect sensitivities/gradients during training.
    \item \textbf{Integrator accuracy}: The intrinsic error associated with the order of the integrator used.
    \end{itemize}
    The first error is practically unavoidable, since without prior knowledge of the Hamiltonian’s functional form, one cannot guarantee that the true Hamiltonian belongs to the chosen approximation space. Consequently, even with perfect data and optimization, an approximation error may remain due to a mismatch between the true model and the representational capacity of the hypothesis class. We can, however, tackle the second and third sources of error. In Section \ref{ss:adjoint_sens}, we described how a symplectic discretization of the continuous adjoint yields the same sensitivities as backpropagation through the discrete solver, so the learning error is not degraded by inconsistent gradients; the rapid, stable convergence of the training curves in Figures \ref{fig:double_well}--\ref{fig:kepler_problem} is consistent with this. The third source of error is due to the intrinsic accuracy of a symplectic integrator. In theory, we can use a very high-order symplectic integrator to reduce that error; however, that comes at a high computational cost. Instead, we can use concepts from geometric integration and reduce the error of our integrator in a completely offline manner without doing any additional ODE solves. We show in the subsequent section that this is possible using the modified Hamiltonian construction. 

\subsection{Modified Hamiltonian and correction of learned Hamiltonian}\label{ss:modified_hamiltonian}
\medskip
In this section, we will validate the claims in Section \ref{ss:advantage_symp} and construct a modified Hamiltonian from the learned Hamiltonian, which is a more accurate representation of the true Hamiltonian. To analyze the errors due to integrator in more detail, in this section, we will assume a noiseless setting unlike in the previous section. Let us establish some terminology.

\begin{itemize}
    \item $\mathbf{\Hc_{true}}$: The true Hamiltonian that we wish to learn. Ideally, it should represent a true physical system from which our training trajectories $(q, p)$ are generated.
    \item ${\Phi_h}$: The one-step numerical map which we use in the forward pass to generate the trajectories.
    \item $\mathbf{\Hc_{NN}}$: The learned Hamiltonian, represented by a neural network.
    \item $\mathbf{\tilde{\Hc}_{NN}}$ : The modified Hamiltonian of the learned Hamiltonian, the exact flow of which agrees with the numerical flow of the learned Hamiltonian, i.e., $\Phi_h^{\Hc_{NN}}=\varphi_h^{\tilde{\Hc}_{NN}}$.
\end{itemize}

\begin{figure}
    \centering
    \includegraphics[width=0.8\linewidth]{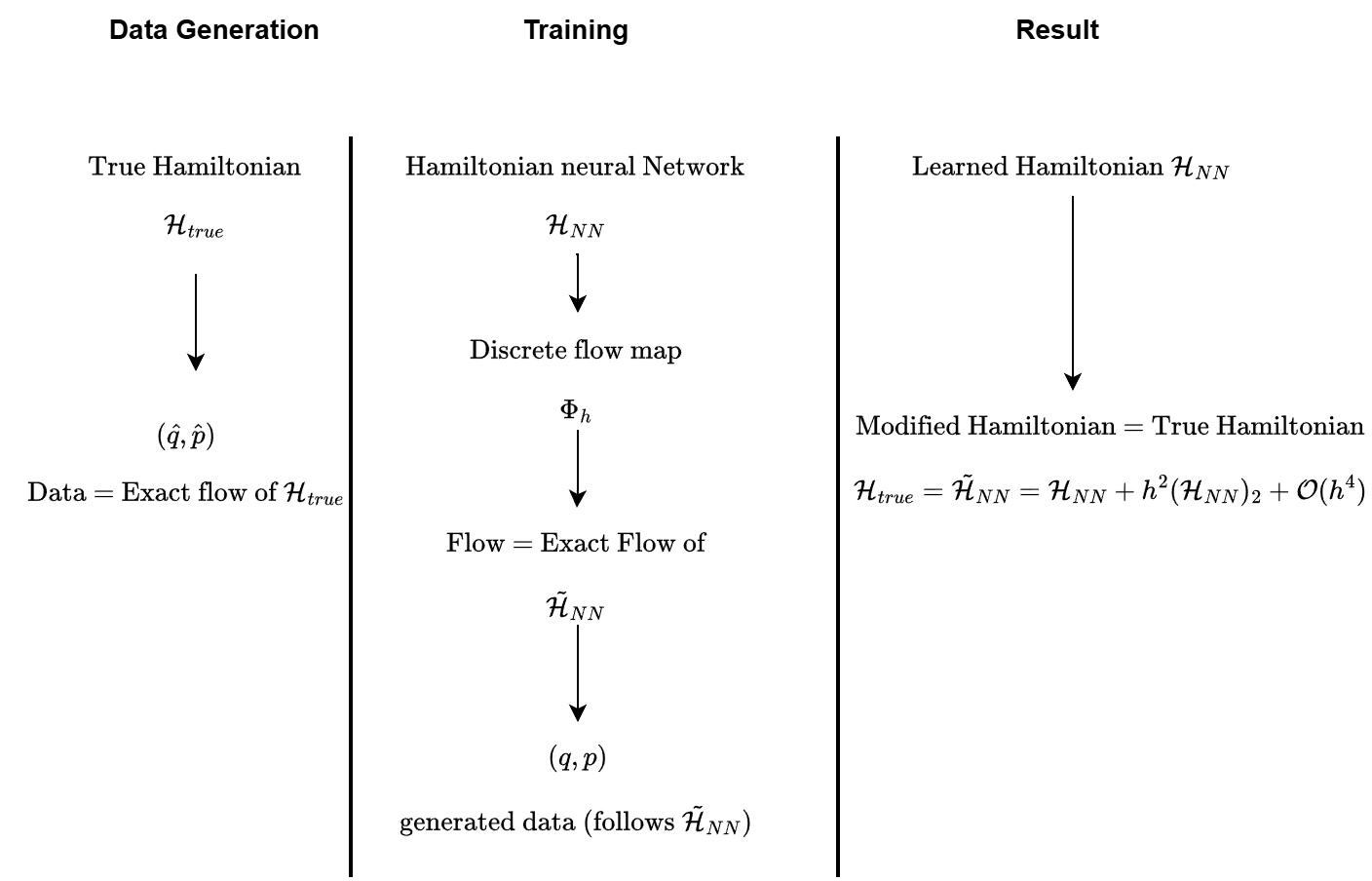}
    \caption{The Hamiltonian learning process for a second-order symplectic map}
    \label{fig:modified_H}
\end{figure}

Suppose we use a neural network to train over the data $(q_i, p_i)$  to learn a Hamiltonian $\Hc_{true}$. Let us assume that the data is noiseless and that the learning is perfect, then the symplectic integrator applied to the learned Hamiltonian $\Hc_{NN}$ agrees with the exact flow of the true Hamiltonian $\Hc_{true}$. By definition, the modified Hamiltonian $\tilde{\Hc}_{NN}$ of the learned Hamiltonian $\Hc_{NN}$ is such that the exact flow of the modified Hamiltonian $\tilde{\Hc}_{NN}$ agrees with the symplectic integrator applied to the learned Hamiltonian $\Hc_{NN}$. From this, we conclude that the modified Hamiltonian of the learned Hamiltonian is the true Hamiltonian that we set out to learn in the first place, as seen in Figure~\ref{fig:modified_H}. Our neural network model aims to learn the Hamiltonian which, when passed through our one-step method $\Phi_h$, generates the true flow. 

Consider a simple Hamiltonian system of a harmonic oscillator given by
\begin{equation}\label{eq:harmonic_oscillator}
    \begin{aligned}
        \Hc_{true} = \frac{1}{2}(p^2 + q^2).
    \end{aligned}
\end{equation}
We use a high-order symplectic integrator (4th order in this case) to generate the trajectory samples $(\hat{q}, \hat{p})$. To isolate the error introduced by the numerical integrator, we assume that the approximation space is sufficiently expressive to contain the true Hamiltonian.
Hence, we consider a simple 2nd degree polynomial ansatz for our neural network $\Hc_{NN}$, given by:
\begin{align*}
    \Hc_{NN} = \theta_0 + \theta_1q + \theta_2p + \theta_3q^2 + \theta_4qp + \theta_5p^2.
\end{align*}
We wish to learn the coefficients $\theta_i$. We use a second-order symplectic map $\Phi_h$ in the forward pass of our neural net, which generates the flow $(q_h, p_h)$. We use a stepsize $h=0.1$ for the forward map. Now we formulate the loss function over the true and predicted flow and optimize according to the scheme described above.
\subsubsection{Modified Hamiltonian}\label{sec:modified_Hamiltonian}
For the implicit midpoint rule, the modified Hamiltonian for the learned Hamiltonian $\Hc_{NN}$ is given using \eqref{eq:modified_Hamiltonian_IM} by
\begin{equation}\label{eq:NN_modified_Hamiltonian}
    \begin{aligned}
        \tilde{\Hc}_{NN} = \Hc_{NN} - \frac{h^2}{24} \nabla^2\Hc_{NN}(J^{-1}\nabla \Hc_{NN}, J^{-1}\nabla \Hc_{NN}) + \mathcal{O}(h^4),
    \end{aligned}
\end{equation}
where the term $\nabla\Hc_{NN}$ is a vector of gradients w.r.t. $q$ and $p$.
Now, from the training, the learned Hamiltonian is
\begin{align*}
   \Hc_{NN} = 0.15410 + (1.1111\times10^{-7})q  -(6.1064\times10^{-8})p +  0.500420q^2 -(1.5463\times 10^{-8})qp +
         0.500415p^2.
\end{align*}
We can immediately observe that the coefficients of $q^2$ and $p^2$ are not the same as in \eqref{eq:harmonic_oscillator}, also we note that there is a non-zero offset which is due to the fact that while training, our symplectic map \eqref{eq:im_int} only sees the derivative of $\Hc$ w.r.t. $q$ and $p$ and hence any Hamiltonian function plus a constant also produces the same dynamics. Neglecting the numerically insignificant terms and correcting for the offset, we can write this as
\begin{equation}\label{eq:learned_H}
    \begin{aligned}
        \Hc_{NN} = 0.500420q^2 + 0.500415p^2.
    \end{aligned}
\end{equation}
 Now, we plug this into the expression for the modified Hamiltonian \eqref{eq:NN_modified_Hamiltonian}, where, for a Hamiltonian of the form $\Hc_{NN}=aq^2+bp^2$, the correction term evaluates to $\nabla^2\Hc_{NN}(J^{-1}\nabla \Hc_{NN}, J^{-1}\nabla \Hc_{NN}) = 8a^2b\,q^2 + 8ab^2\,p^2$, giving

 \begin{align*}
     \tilde{\Hc} = 0.500420q^2 + 0.500415p^2 - \frac{(0.1)^2}{24} (1.00251q^2 + 1.00250p^2) + \mathcal{O}(h^4).
 \end{align*}
 
\begin{figure}[!ht]
    \centering
    \begin{subfigure}{0.40\textwidth}
        \includegraphics[width=\textwidth]{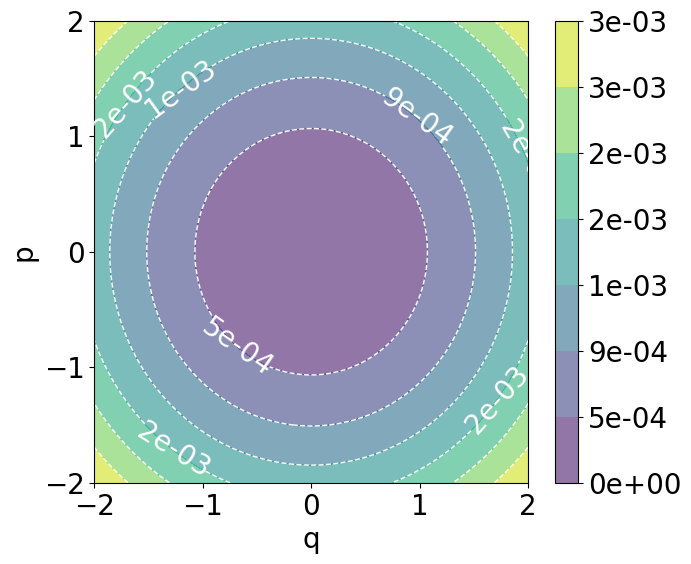}
        \caption{}
        \label{fig:subfig1_h_vs_htrue}
    \end{subfigure}
    \hspace{1.0cm}
    \begin{subfigure}{0.40\textwidth}
        \includegraphics[width=\textwidth]{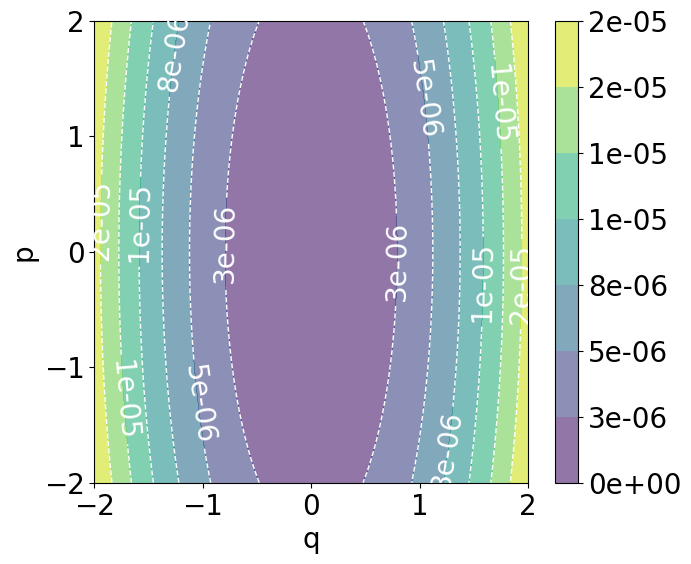}
        \caption{}
        \label{fig:subfig2_hmod_vs_htrue}
    \end{subfigure}
    
    \caption{\textbf{Left} The error contours for $|\Hc_{true} - \Hc_{NN}|$ and \textbf{Right} $|\Hc_{true} - \tilde{\Hc}_{NN}|$ where the modified Hamiltonian is written up to fourth-order (contours now show the $\mathcal{O}(h^4)$ errors). We can see that the modified Hamiltonian of the learned Hamiltonian is a better approximation of the true Hamiltonian than the learned Hamiltonian.}
\label{fig:true_vs_modified_H}
 \end{figure}

 Neglecting the fourth-order terms, we
 obtain the modified Hamiltonian for our learned Hamiltonian,
 \begin{equation}\label{learned_modified_H}
 \begin{aligned}
 \tilde{\Hc}_{NN} \approx 0.500002q^2 + 0.499997p^2,
 \end{aligned}
 \end{equation}
 where we have neglected the fourth-order terms and we observe that it agrees better with the true Hamiltonian, as seen in Figure \ref{fig:true_vs_modified_H}. In this manner, even with a low-order integrator and post-processing, via constructing a modified Hamiltonian, we can get a better approximation of the true Hamiltonian without any additional computation cost. In this example, we constructed the modified Hamiltonian up to second order. One can do it for higher orders and can get an increasingly better approximation.

For the other examples discussed in Section~\ref{ss:hamiltonian_systems}, it is possible to obtain a more accurate approximation of the true Hamiltonian by post-processing the learned Hamiltonian $\Hc_{NN}(q,p)$. The corresponding results are reported in Table~\ref{tab:baseline_comparison}, together with those of two baseline approaches described in \cite{david2023symplectic, xiong2020nonseparable}. To ensure a fair comparison of both the learning performance and the quality of the underlying numerical integrators, all methods were evaluated under similar experimental settings. Each method used a neural network with one hidden layer containing $100$ neurons and was trained until the validation loss had sufficiently converged. The training batch size was fixed to the default choice recommended for each baseline, while a batch size of $512$ was used for both training and validation in our method. The ground-truth trajectories were generated at a temporal resolution of $\Delta t=0.01$ using a fourth-order symplectic integrator. The same $32{,}768$ training trajectories were used for all methods, with initial conditions sampled uniformly from $\Omega^{d}$, where $\Omega$ denotes the corresponding phase-space domain and $d$ is the dimension of the system. Noiseless training data were used in this comparison to isolate the effect of the modified-Hamiltonian construction. For every method, the training loop used a step size of $h=0.1$. The prediction error was evaluated using both the average $L_1$ error and the $L_\infty$ error over a uniform Cartesian grid. For systems of dimension greater than two, two phase-space coordinates were varied over the grid while the remaining coordinates were held fixed to random values in domain, thereby restricting the evaluation to a two-dimensional subspace. The modified Hamiltonian was constructed using Equation~\eqref{eq:NN_modified_Hamiltonian}, where the gradient was computed using automatic differentiation and the Hessian was assembled componentwise by differentiating each component of the gradient. The results show that for a fixed model size, our method performs significantly better compared to the other two methods at test. The first method uses an extended phase space to make the system explicit, whereas the second method does not solve the implicit system but rather matches the gradient at the midpoint. The results in the last two columns show that the modified Hamiltonian systematically reduces both the average $L_1$ error and the $L_\infty$ error, although the improvement is modest: roughly one percent for the double well, progressively smaller for the coupled oscillator and H\'enon--Heiles systems, and below the displayed precision for the Kepler problem. This is consistent with the error decomposition of Section~\ref{ss:sources_of_error}: for a neural-network ansatz, the approximation error dominates the $\mathcal{O}(h^2)$ integrator error that the modified-Hamiltonian correction removes, so the correction is only marginally visible. In contrast, for the polynomial ansatz above, whose hypothesis space contains the true Hamiltonian, the same correction reduces the coefficient error by two orders of magnitude.

\begin{table*}[t]
  \centering
  \small
  \setlength{\tabcolsep}{6pt}
  \renewcommand{\arraystretch}{1.15}
  \caption{ The average $L_1$ error and the $L_\infty$ error in Hamiltonian prediction. For two-dimensional systems, the errors are evaluated over points sampled on a uniform Cartesian grid covering the phase-space domain $\Omega$. For systems with dimension greater than two, two phase-space coordinates are varied over a uniform Cartesian grid while the remaining coordinates are held fixed, thereby restricting the evaluation to a two-dimensional subspace. The two columns under \emph{This Work} report the errors for the learned Hamiltonian $\mathcal{H}_{NN}$ and the modified Hamiltonian $\widetilde{\mathcal{H}}_{NN}$ relative to the true Hamiltonian $\mathcal{H}_{\mathrm{true}}$, respectively. }
  \label{tab:baseline_comparison}

  \begin{tabular}{@{}llcc|cc@{}}
    \toprule
    \multirow{2}{*}{System}
    & \multirow{2}{*}{Metric}
    & \multirow{2}{*}{NSSNN \cite{xiong2020nonseparable}}
    & \multirow{2}{*}{SHNN \cite{david2023symplectic}}
    & \multicolumn{2}{c}{This Work} \\
    \cmidrule(lr){5-6}
    &
    &
    &
    &
    $|\mathcal{H}_{NN}
      -\mathcal{H}_{\mathrm{true}}|$
    &
    $|\widetilde{\mathcal{H}}_{NN}
      -\mathcal{H}_{\mathrm{true}}|$ \\
    \midrule

    \multirow{2}{*}{DW}
    & $\langle L_1\rangle$
    & $7.0358\times 10^{-1}$
    & $4.0685\times 10^{-1}$
    & $4.5493\times 10^{-3}$
    & $4.5044\times 10^{-3}$ \\
    & $L_\infty$
    & $3.1367$
    & $1.0242$
    & $2.7993\times10^{-2}$
    & $2.7732\times10^{-2}$ \\
    \midrule

    \multirow{2}{*}{CHO}
    & $\langle L_1\rangle$
    & $8.3887\times 10^{-1}$
    & $1.2190\times 10^{-3}$
    & $6.3517\times 10^{-4}$
    & $6.3280\times 10^{-4}$ \\
    & $L_\infty$
    & $4.6189$
    & $1.5624\times 10^{-2}$
    & $4.8381\times 10^{-3}$
    & $4.8180\times10^{-3}$ \\
    \midrule

    \multirow{2}{*}{HH}
    & $\langle L_1\rangle$
    & $4.6282\times 10^{-2}$
    & $1.7880\times 10^{-2}$
    & $9.1568\times 10^{-4}$
    & $9.1516\times 10^{-4}$ \\
    & $L_\infty$
    & $2.0621\times10^{-1}$
    & $8.2698\times10^{-2}$
    & $2.0308\times10^{-3}$
    & $2.0284\times10^{-3}$ \\
    \midrule

    \multirow{2}{*}{Kepler}
    & $\langle L_1\rangle$
    & $6.0778\times 10^{-1}$
    & $6.2212\times 10^{-1}$
    & $2.7577\times 10^{-1}$
    & $2.7577\times 10^{-1}$ \\
    & $L_\infty$
    & $1.8394$
    & $1.2025$
    & $8.7415\times10^{-1}$
    & $8.7414\times10^{-1}$ \\

    \bottomrule
  \end{tabular}
\end{table*}

\section{Conclusion}
\medskip
The adjoint approach normally results in gradients that differ from backpropagation, unless the adjoint system is computed using the cotangent lift of the numerical integrator used in the forward propagation, in which case the adjoint approach yields gradients that coincide with backpropagation. When the forward flow is integrated using a Runge--Kutta method, the cotangent lift is a symplectic partitioned Runge--Kutta method, where the original Runge--Kutta integrator to the primal variables, and the adjoint variables are integrated by the symplectic adjoint of the original Runge--Kutta integrator, defined by \eqref{eq:symp_condition}.


When the forward flow is Hamiltonian, as is the case for Hamiltonian Neural Networks, it is natural to discretize the forward flow using a symplectic integrator. If the forward integrator is a symplectic Runge--Kutta method, then the symplectic adjoint is itself. As such, using the same symplectic integrator on the adjoint variables will lead to a discretization of the adjoint system that yields gradients that also coincide with backpropagation, leading to an efficient method for training Hamiltonian Neural Networks.

In our work, we adopt such an approach, using implicit symplectic partitioned Runge--Kutta methods. Symplectic methods are generally implicit for non-separable Hamiltonians, unless one artificially doubles the number of variables in an augmented formulation. However, contrary to conventional wisdom, implicit SPRK methods can be very efficiently implemented by using an explicit RK method of the same order as a predictor, and using a few fixed-point iterations of the SPRK method as a corrector. Therefore, Hamiltonian Neural Networks with a non-separable Hamiltonian ansatz can be efficiently trained using implicit SPRK discretization by applying the adjoint method combined with the predictor-corrector fixed-point iteration.

Finally, we observe that the learned Hamiltonian obtained using a lower-order symplectic integrator can be post-processed using backward error analysis to obtain a modified Hamiltonian that better agrees with the true Hamiltonian, without the need to use higher-order integrators in the learning process.

\section*{Acknowledgements}{
\begin{itemize}
    \item HC and VK acknowledge support from the Czech National Science Foundation under Project 24-11664S
    \item HC acknowledges the computational resources provided by RCI project. 
    \item ML was supported in part by NSF under grants CCF-2112665, DMS-2307801, by AFOSR under grant FA9550-23-1-0279.
\end{itemize}}

\section*{Disclaimer}
This paper was prepared for information purposes
and is not a product of HSBC Bank Plc. or its affiliates.
Neither HSBC Bank Plc. nor any of its affiliates make
any explicit or implied representation or warranty and
none of them accept any liability in connection with
this paper, including, but not limited to, the completeness,
accuracy, reliability of information contained herein and
the potential legal, compliance, tax or accounting effects
thereof. Copyright HSBC Group 2025.

\bibliography{hnn_iop}

\pagebreak

\end{document}